\documentclass{bmvc2k}

\usepackage{times}
\usepackage{epsfig}
\usepackage{amsmath}
\usepackage{amssymb}
\usepackage{xspace}
\usepackage{booktabs}
\usepackage{soul}
\usepackage{longfbox}
\usepackage{graphbox,graphicx}
\usepackage{caption}

\newcommand{\MPG}{\textit{MPG}\xspace}
\newcommand{\MSMAE}{\textit{MSMAE}\xspace}
\newcommand{\AR}{\textit{AR}\xspace}
\newcommand{\pizzaIngr}{\textit{Pizza10}\xspace}
\newcommand{\pizzaView}{\textit{PizzaView}\xspace}

\DeclareMathOperator{\x}{\mathbf{x}}
\DeclareMathOperator{\y}{\mathbf{y}}
\DeclareMathOperator{\z}{\mathbf{z}}
\DeclareMathOperator{\te}{\mathbf{t}}
\DeclareMathOperator{\w}{\mathbf{w}}

\title{Multi-attribute Pizza Generator: Cross-domain Attribute Control with Conditional StyleGAN}

\addauthor{Fangda Han}{fh199@cs.rutgers.edu}{1}
\addauthor{Guoyao Hao}{gh343@scarletmail.rutgers.edu}{1}
\addauthor{Ricardo Guerrero}{r.guerrero@samsung.com}{2}
\addauthor{Vladimir Pavlovic}{vladimir@cs.rutgers.edu}{1}

\addinstitution{
    Rutgers University\\
    Piscataway, NJ, USA
}
\addinstitution{
    Samsung AI Center\\
    Cambridge, UK
}

\runninghead{F. Han, G. Hao, R. Guerrero, V. Pavlovic}{Multi-attribute Pizza Generator}

\def\eg{\emph{e.g}\bmvaOneDot}

\begin{document}

\maketitle

\vspace{-1em}
\begin{abstract}
Multi-attribute conditional image generation is a challenging problem in computer vision.
We propose \textbf{Multi-attribute Pizza Generator (\MPG)}, a conditional Generative Neural Network (GAN) framework for synthesizing images from a trichotomy of attributes: content, view-geometry, and implicit visual style.
We design \MPG by extending the state-of-the-art StyleGAN2, using a new conditioning technique that guides the intermediate feature maps to learn multi-scale multi-attribute entangled representations of controlling attributes. 
Because of the complex nature of the multi-attribute image generation problem, we regularize the image generation by predicting the explicit conditioning attributes (ingredients and view).
To synthesize a pizza image with view attributes outside the range of natural training images, we design a CGI pizza dataset \textbf{\pizzaView} using 3D pizza models and employ it to train a view attribute regressor to regularize the generation process, bridging the real and CGI training datasets.
To verify the efficacy of \MPG, we test it on \textbf{\pizzaIngr}, a carefully annotated multi-ingredient pizza image dataset.
\MPG can successfully generate photorealistic pizza images with desired ingredients and view attributes, beyond the range of those observed in real-world training data. 
\end{abstract}

\section{Introduction}
\vspace{-0.5em}

A meal appearing in a food image is a product of a complex progression of cooking stages, where multiple ingredients are combined using mechanical and thermo-chemical actions (cooking techniques), followed by physical and stylistic arrangements of food on a plate (plating).  Food images are, thus, naturally diverse, displaying complex and highly variable entanglement of shapes, colors, and textures.  For instance, pepperoni on a pizza is typically round, but its final appearance depends on the cooking time and cooking temperature.  Multiple ingredients mixed together will overlay and obscure each other.  A pizza can be rectangular or oval, a  pie or a single slice, pictured whole or zoomed in for perceptual or artistic impact.  While 
Generative Adversarial Networks~\cite{goodfellow2014generative} (GANs) have been used to create high-fidelity images~\cite{karras2019style, karras2020training, brock2018large} of complex structured objects, such as faces, bodies, birds, cars, etc., they remain challenged by domains such as the food images.  

Previous works on food image generation either follow the image-to-image translation paradigm~\cite{ito2018food, papadopoulos19cvpr} or generate blurry images~\cite{han2020cookgan, zhu2020cookgan} as a function of a single type of attributes (text of ingredients/instructions).  In this work, we propose a new way to extend the capacity of GAN models to solve the problem of generating food images conditioned on the trichotomy of attributes: food content (ingredients), geometric style (view point, scale, horizontal and vertical shift), and visual style (diversity in fine-grained visual appearance of ingredients and the final dish).  A main challenge we tackle here is that, for explicit control of different food attributes, no single real food image dataset is sufficiently diverse or fully attribute labeled, to allow learning of photo-realistic conditional attribute models.   

To this end, we propose a conditional GAN called Multi-attribute Pizza Generator (\MPG) inspired by~\cite{karras2020analyzing}. 
\MPG first encodes a set of conditioning attributes, ingredients and geometric view parameters, into entangled scale-specific embeddings, which are then injected into the synthesis network together with the visual ``style noise'' to create images.  
To address the lack of simultaneous (paired) labels for food content and geometric style in a single dataset, the \MPG leverages integration of two new datasets with individual (unpaired) attribute category labels: \pizzaIngr, a real-world pizza dataset with ingredient annotations, and \pizzaView, a CGI dataset created from 3D pizza models annotated with geometric style attributes. 
\pizzaView is used to train a geometric style attribute regressor, which subsequently regularizes the conditional generator, trained on \pizzaIngr, to generate natural food images with arbitrary view attributes. 
The implicit visual style is learned in an unsupervised manner.

Our main contributions are: 
(1) We propose a Multi-attribute conditional GAN framework, which extends StyleGAN2~\cite{karras2020analyzing}.  The framework incorporates a Multi-scale Multi-attribute Encoder (\MSMAE) and Attribute Regularizers (\AR{}s) that guide generator to synthesize the desired attributes.
(2) Starting from pizzaGANdata~\cite{papadopoulos19cvpr}, we create a real-world dataset \pizzaIngr, by relabeling a subset of consistent ingredients. \pizzaIngr opens up the possibility of learning high-performing multi-ingredient prediction models by eliminating the ubiquitous label noise of~\cite{papadopoulos19cvpr}.
(3) We design a CGI dataset \pizzaView using 3D pizza models with controlled geometric style attributes (camera angle, scale, x-shift and y-shift), used to train the view attribute regressor\footnote{Code and \pizzaView dataset: https://github.com/klory/MPG2.}.
Together, these contributions make it possible to learn photo-realistic food image synthesis models, from disparate (real \& CGI) domain attribute data, able to generate images outside the range of natural training image attributes.

The reminder of this manuscript is organized as follows: \autoref{sec:related_works} introduces the related works including GAN, conditional GAN, food generation and multi-label image generation. 
\autoref{sec:methodology} presents our conditional StyleGAN framework \MPG as well as the Multi-Scale Multi-Attribute Encoder (\MSMAE) and Attribute Regularizer (\AR). 
\autoref{sec:pizza10} and \autoref{sec:pizzaView} describe the motivation and methodology to create \pizzaIngr and \pizzaView datasets.
\autoref{sec:experiments} includes experiments that verify the effectiveness of our framework, an ablation study, and an assessment of generated images.
Finally, in \autoref{sec:conclusion} we summarize our conclusions.

\section{Related Works}
\vspace{-0.5em}
\label{sec:related_works}

\noindent \textbf{GANs}~\cite{goodfellow2014generative} are trained as a two-player min-max game to estimate the true data distribution from samples. 
In the past few years, GANs have evolved rapidly, improving their ability to generate high-fidelity digital images of compact objects and natural scenes~\cite{miyato2018spectral, brock2018large, karras2020training}, as well as human faces~\cite{karras2019style, karras2020analyzing}.

\vspace{0.2em}
\noindent \textbf{Conditional GAN}~\cite{mirza2014conditional} extends GAN by conditioning generation process with auxiliary information to control the appearance of generated images, \eg controlling handwriting digit in MNIST~\cite{mirza2014conditional}, natural object or scene category in ImageNet dataset~\cite{miyato2018cgans}, or the category of CIFAR10 dataset~\cite{karras2020training}.
\cite{oeldorf2019loganv2} applies StyleGAN to the logo generation task, our task differs from this as we seek to condition on more than a single label (i.e., discrete ingredients and continuous view attributes); thus, the structures and spatial relationships among labels are far more diverse.
GAN-Control~\cite{shoshan2021gancontrol} also utilizes StyleGAN to control different attributes (\eg human age, pose, illumination, expression and hair color), but they need to split the mapping network into different sub networks for each attribute and construct mini-batch deliberately during training, they also need a second phase to fit individual attribute encoders between latent representations and label per attribute. 
Our method controls the attributes by encoding the attribute values directly and the model needs to be trained only in a single phase.

Another line of research assumes conditioning in the form of another image (i.e., image-to-image generation such as image editing and/or image inpainting), with the goal of changing its attributes~\cite{he2019attgan, choi2018stargan} or associated styles~\cite{isola2017image, zhu2017unpaired}. 
Note the task of our proposed work is complex since different pizzsa often contains different ingredients in terms of shape, color and amount, and we only leverage attributes \& labels as the conditioning input, lacking the strong input image prior.
Finally, image GANs can be conditioned directly on natural language textual descriptions~\cite{zhang2018stackgan++, xu2018attngan, li2019controllable}. 
While, in principle, more general, these approaches in practice work well on objects such as birds or flowers that share similar structures within classes, but tend to fail on complex scene generation, like those in COCO dataset.

\vspace{0.2em}
\noindent \textbf{Food Image Generation} is usually accomplished using a conditional GAN to attain higher fidelity.
\cite{ito2018food, papadopoulos19cvpr} apply GANs to transform the food type or add/remove ingredients from a source image.  This yields food images of reasonable quality, but requires image-based conditioning.  In text-based conditioning,
\cite{sabinigan} concatenate the embedding from a pretrained text encoder with noise to generate a food image.
Other approaches, c.f., \cite{el2019gilt}, \cite{han2020cookgan}, or \cite{wang2019food}, employ more sophisticated generation models such as  StackGAN2~\cite{zhang2018stackgan++} or ProgressiveGAN~\cite{karras2017progressive}.
However, these works are only able to generate low-resolution images of food-like texture, but with indistinguishable ingredients and missing details. They also lack comprehensive evaluation metrics, such as the input attribute reconstruction ability in addition to image appearance metrics, like the FID, that provide complementary assessment needed to appraise image realism. Moreover, these methods are unable to control other important image aspects, such as the geometric style.  In contrast, our framework is able to synthesize images of high fidelity, with distinct detail in the form of more naturally appearing ingredient elements. We couple this with the ability to explicitly control the view geometry, exceeding the range of view parameters in natural food image datasets.  Our extensive evaluation, including using the new conditional FID and mAP scores, affirms this judgement.

\vspace{0.2em}
\noindent \textbf{Text-based Image Generation}
\cite{zhang2018stackgan++} uses a stacked multi-scale generator to synthesize images based on label embeddings.
~\cite{xu2018attngan} extends~\cite{zhang2018stackgan++} and synthesize details at different subregions of the image by paying attentions to the relevant words in the natural language description. 
\cite{qiao2019mirrorgan} extends \cite{xu2018attngan} by regularizing the generator and redescribing the corresponding text from the synthetic image. 
While free text is a potentially stronger signal, it is often corrupted by image-unrelated signals, such as writing style. 
On the other hand, while image labels may appear more specific than free-text, the visual diversity of objects described by those labels, such as food ingredients, and changes in appearance that arise from interactions of those objects can make the multi-label image generation problem an equal if not greater challenge to that of the caption-to-image task. 
\vspace{-1em}

\section{Methodology}
\label{sec:methodology}
\vspace{-0.5em}

\begin{figure*}[t!]
    \centering
    \begin{minipage}[t]{0.48\linewidth}
        \centering
        \includegraphics[align=t,width=\textwidth]{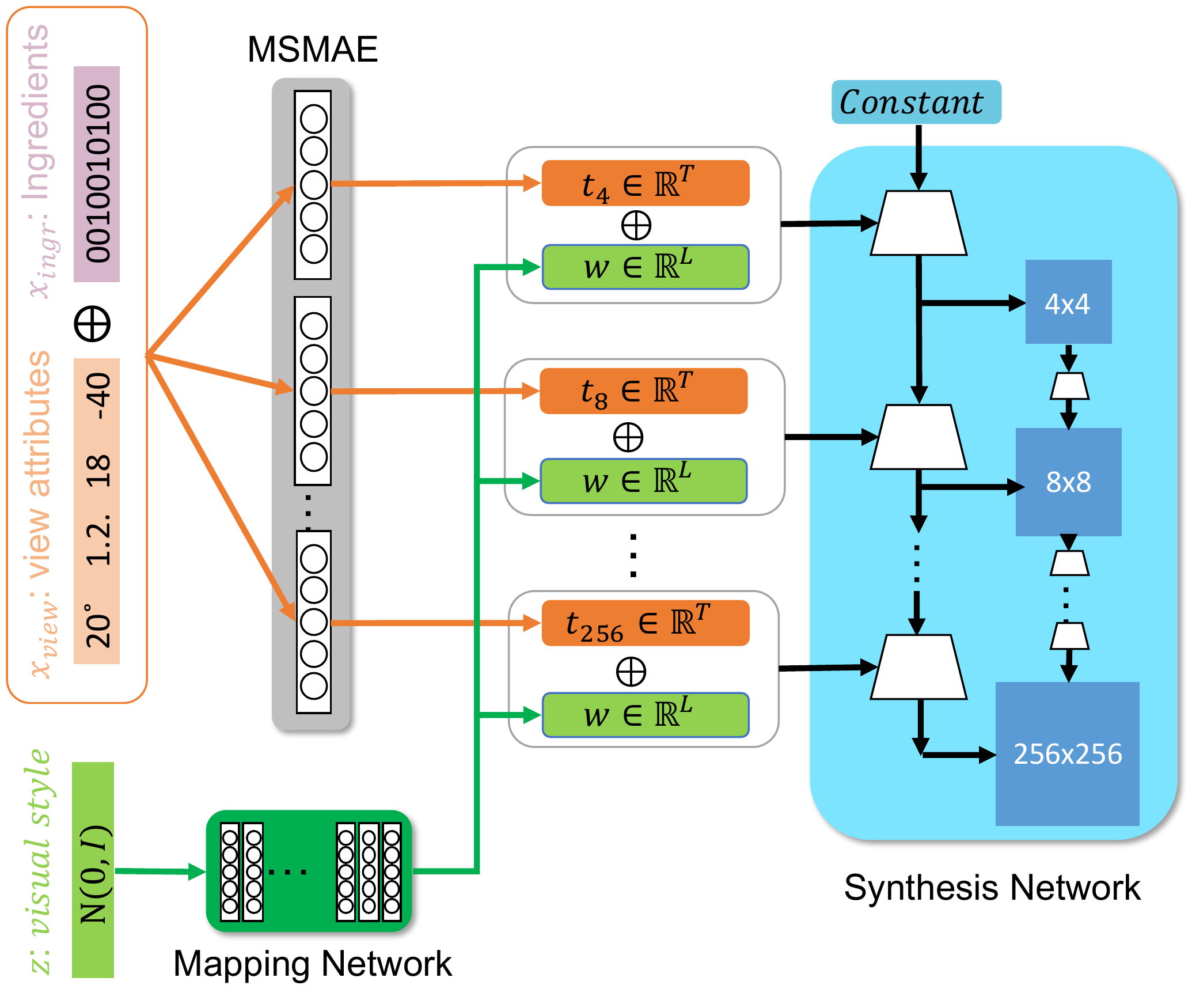}
        \caption*{\footnotesize (a) Generator. \textbf{Top-Left}: Multi-Scale Multi-Attribute Encoder (\MSMAE), each scale has its own layers. Ingredients are represented by a binary vector and then concatenated with view attributes (view angle, scale, x-shift, y-shift). \textbf{Bottom-Left}: Mapping Network to encode visual style. \textbf{Right}: Synthesis Network. $\oplus$ means concatenation, notice images at different scales are conditioned with different label embedding}
    \end{minipage}
    \hfill
    \begin{minipage}[t]{0.48\linewidth}
        \includegraphics[align=t,width=\textwidth]{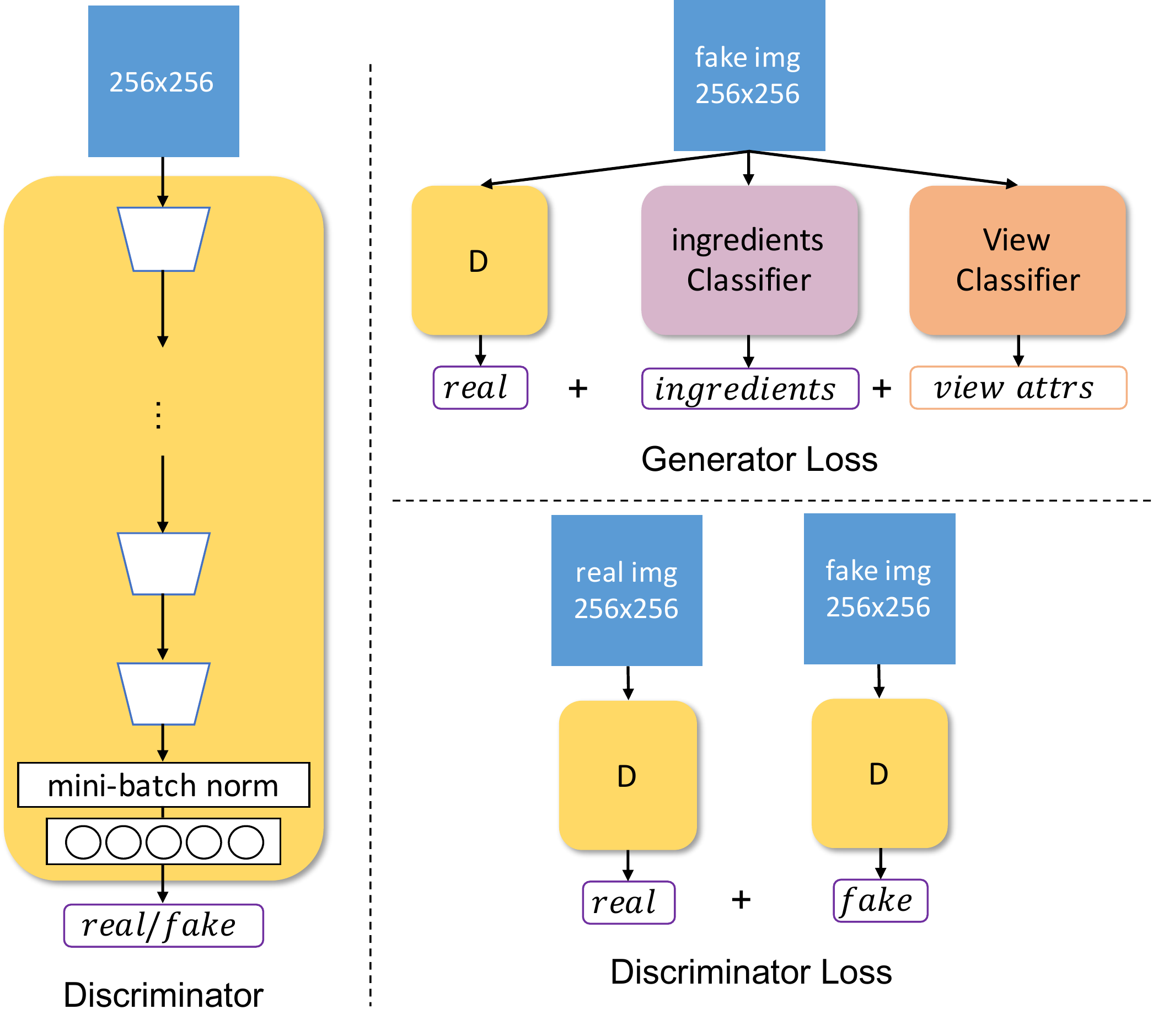}
        \caption*{\footnotesize (b) \textbf{Left}: Discriminator. \textbf{Top right}: Generator Loss, consisting of discriminator loss, ingredients classification loss, and view regressor loss. \textbf{Bottom right}: Discriminator Loss, consisting of discriminator losses for real image and fake image.}
    \end{minipage}
    \vspace{0.5em}
    \caption{Framework overview}
    \vspace{-1em}
    \label{fig:mpg_framework}
\end{figure*}

We adjust StyleGAN2 and create Multi-attribute Pizza Generator (\MPG), making it possible to generate pizza images with specific, controllable attributes (including ingredients and views) and combinations thereof.

Formally, given a vector $\x$ indicating the attribute list (\autoref{fig:mpg_framework}(a) Left) and some random visual style $\z$, we seek to generate a pizza image $\y$ that contains the desired attributes. We propose to use a model endowed with Multi-Scale Multi-Attribute Encoder (\MSMAE) and a Attribute Regularizer (\AR) to solve this problem.

\subsection{Multi-Scale Multi-Attribute Encoder}
StyleGAN2 transforms and injects style noise $\z$ into the generator at each scale. 
Inspired by this, we propose \MSMAE (Multi-Scale Multi-Attribute Encoder in \autoref{fig:mpg_framework}(a) top-left) to condition the synthesis network on the attributes $\x = \x_{ingr} \oplus \x_{view}$, where $\x_{ingr}$ is the ingredients binary vector, and $\x_{view}$ represents view attributes (i.e., camera angle, scale, x-shift, y-shift).

\MSMAE consists of a set of sub-encoders $\{Enc_i(\x)\}$ (one for each scale $i$) and takes $\x$ as input to output label embedding $\{\te_i\}$ (also one for each scale $i$). Formally, if the generated image $\tilde{\y}$ is of size $256^2$,
\begin{align}
    \{\te_i\} = \text{MSMAE}(\x) = \{Enc_i(\x)\} \, ,i \in \{4,8,...,256\}.
    \label{eq:msmae}
\end{align}

We implement $Enc_i$ as a multi-layer perceptron. The mapping network $f$ is kept as in StyleGAN2 to provide diversity to the output image. 
The label embedding $\{\te_i\}$ is then concatenated with the output from the mapping network together as the input of the synthesis network $S$. 
We inject $\te_i\oplus\w$ the same way as in StyleGAN2 and generate the output image $\tilde{\y}$
\begin{align}
    \tilde{\y} = G(\x, \z) = S\left(\{ \te_i \oplus \w \} \right) = S\left(\{\text{MSMAE}({\x}) \oplus f(\z) \} \right),
\end{align}
where G is the generator composed of $\{\text{MSMAE}, f, S\}$.

\subsection{Attribute Regularizer}
To generate pizza image with desired attributes, we use two pretrained model (ingredients classifier $h_{ingr}$ and view attribute regressor $h_{view}$ to regularize training process. An important aspect to note about the \pizzaIngr dataset, used to train \MPG, is that it does not contain view attribute labels, therefore, during training these are imputed with predicted values from the view attribute regressor $h_{view}$, same which is trained exclusively on the CGI \pizzaView dataset. Refer to \autoref{sec:pizzaView} for more details about the performance of $h_{view}$.

\vspace{0.2em}
\noindent\textbf{Ingredient Attribute Regularizer.} Inspired by~\cite{odena2017conditional, han2020cookgan}, we pretrain a multi-ingredient classifier $h_{ingr}$ on pairs of real image and its ingredient list $\{(\y, \x)\}$ with binary cross entropy loss.
Then during training the generator, we regularize the generated image $\tilde{\y}$ to predict the correct ingredients $\x_{ingr}$. The Attribute Regularizer for ingredients here can be formalized as
\begin{align}
    \mathcal{L}_{ingr} = \text{BCE} \left( h_{ingr}(\tilde{\y}), \x_{ingr} \right).
    \label{eq:ingr_loss}
\end{align}

\vspace{0.2em}
\noindent\textbf{View Attribute Regularizer.} We apply view attribute regressor $h_{view}$ similarly using L2 loss to regularize the generator as well. 
Given the view attributes $\x_{view}$~\footnote{Actually, $x_{view}$ is unknown because real images don't contain view labels, we instead use $h_{view}$ trained on \pizzaView to predict the view attributes as pseudo label $\hat{\x}_{view}$, please read \autoref{sec:pizzaView} for more details.} and the synthesized image $\tilde{\y}$, the loss can be formalized as
\begin{align}
    \mathcal{L}_{view} = \text{L2} \left( h_{view}(\tilde{\y}), \x_{view} \right).
    \label{eq:view_loss}
\end{align}

\subsection{Loss}
The generator's objective consists of GAN loss and two attribute regularizations
\begin{align}
    \mathcal{L}_G = -D( G(\x, \z) ) + \lambda_{ingr} \mathcal{L}_{ingr} + \lambda_{view} \mathcal{L}_{view},
\label{eq:g_loss}
\end{align}
where $\lambda_{ingr}$ and $\lambda_{view}$ are the weights of two attribute regularization terms.

The discriminator's objective consists of GAN loss with R1 regularization~\cite{drucker1992improving} applied to penalize the gradient flow from the image to the output, shown to stabilize GAN training, which can be formalized as
\begin{align}
    \mathcal{L}_D = D(G(\x, \z)) - D(\y) + \lambda_{r1} \dfrac{\partial D(\x)}{\partial \x},
\label{eq:d_loss}
\end{align}
where $\lambda_{r1}$ is the weight of R1 regularizer.

\section{Pizza10 Dataset}
\label{sec:pizza10}
\vspace{-0.5em}

The performance of neural network models critically depends on the quality of the dataset used to train the models.
We begin with pizzaGANdata~\cite{papadopoulos19cvpr}, which contains $9213$ pizza images annotated with 13 ingredients.
We manually inspected the labeling in this dataset and found a significant portion of it to be mislabeled.
To address this issue, using a custom-created collaborative web platform, we relabeled the dataset and removed three ingredients (\textit{Spinach, Pineapple, Broccoli}) with fewer number of instances (\textit{Corn} is kept since the distinct color and shape makes it a recognizable ingredient even with fewer samples than \textit{Spinach}). 
The refined dataset, \pizzaIngr, decreases the number of `empty samples' (pizza images with no ingredients labeled) by $40\%$ and increases the number of samples for each ingredient category.
We verify the impact of the curated \pizzaIngr on the Multi-ingredient Classification task: given a pizza image, we seek to predict its ingredients.
We initialized the classifier from ResNet50~\cite{he2016deep} with the final fully-connected layer changed to match the output dimension ($\mathbb{R}^{13}$ for pizzaGANdata and $\mathbb{R}^{10}$ for \pizzaIngr). 
The model is trained using binary cross entropy loss, $80\%$ data are used in training while $20\%$ data are for testing.
After training, we use the mean average precision (mAP) as the metric.

\begin{figure}[t]
    \centering
    \begin{minipage}[t]{0.48\linewidth}
        \centering
        \includegraphics[width=\textwidth]{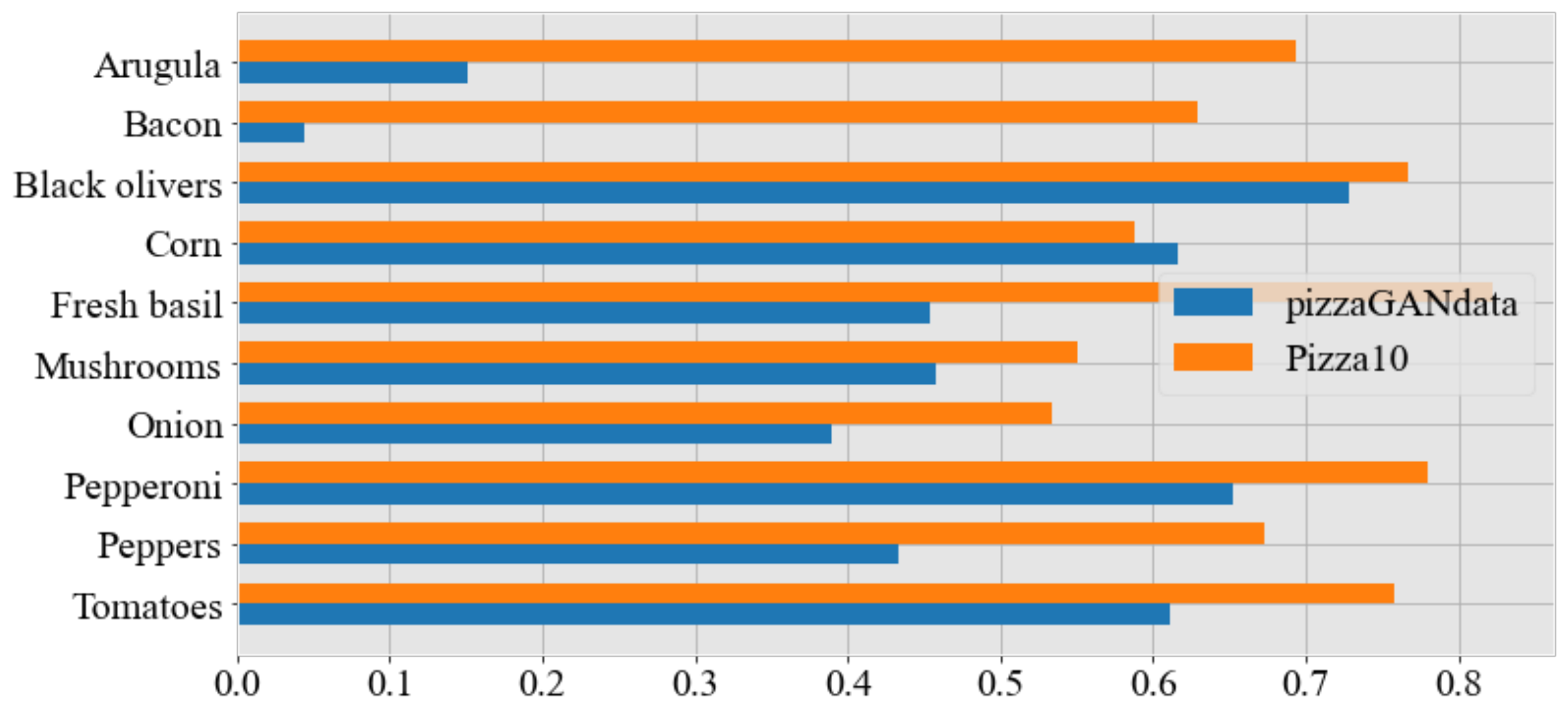}
        \caption*{(a)}
    \end{minipage}
    \hfill
    \begin{minipage}[t]{0.48\linewidth}
        \centering
        \includegraphics[width=\textwidth]{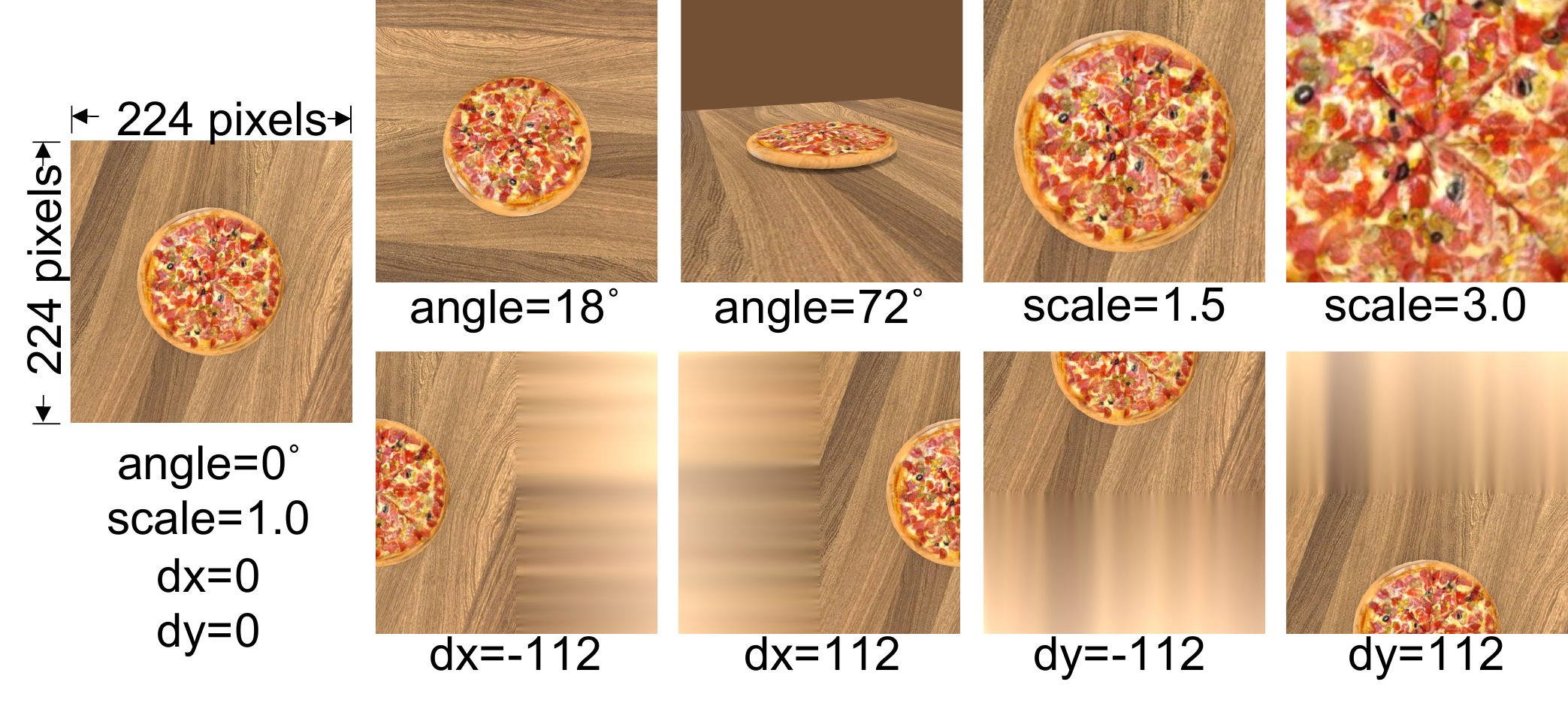}
        \caption*{(b)}
    \end{minipage}
    \vspace{0.5em}
    \caption{(a) Average precision comparison between pizzaGANdata~\cite{papadopoulos19cvpr} and \pizzaIngr dataset. (b) Example images with different view attributes. Left: Reference image. Right: image transformation along each view attribute}
    \vspace{-1em}
    \label{fig:pizza10_performance_and_pizzaView_examples}
\end{figure}

The results are shown in \autoref{fig:pizza10_performance_and_pizzaView_examples}(a). The same model displays a large performance improvement on the refined \pizzaIngr dataset. Upon close inspection, the average precision (AP) for each ingredient increases on most ingredients except \textit{Corn} for the classifier trained on \pizzaIngr, therefore demonstrating the importance of careful curating.
\vspace{-1em}

\section{PizzaView Dataset}
\label{sec:pizzaView}
\vspace{-0.5em}

We also aim to control the view attributes of synthesized images besides ingredients.
However, \pizzaIngr does not contain view attributes; to train an effective view attribute regressor, we create \pizzaView from CGI rendered 3D pizza models. 

There are two advantages to creating a new pizza dataset from 3D models compared with directly labelling the view attributes in \pizzaIngr: (1) The view attributes in \pizzaIngr are  unevenly distributed, with pizzas usually assuming a center image position, with a small range of scales and viewpoints; these narrow-support attribute distributions will make the training of comprehensive, wide-support view attribute regressors difficult. (2) Manual labeling is laborious and prone to errors, with the labelling process further complicated by ambiguities, such as when the pizza is only partially visible, resulting in high label noise or bias.

To construct \pizzaView, we first download $40$ different 3D pizza models from TurboSquid\footnote{https://www.turbosquid.com/}, then import these models into Blender\footnote{https://www.blender.org/} and use Blender's Python API to extract pizza images at different view angles. Without loss of generality, we define the camera viewpoint above the center of the pizza as $0^{\circ}$, then move the camera between $[0^{\circ}$, $75^{\circ}]$ on the view hemisphere to generate images with different \textit{view} attributes (we choose $75^{\circ}$ as the natural extreme range of pizzas in \pizzaIngr). 
After rendering the 2D images with various camera angle from 3D models, we further transform the 2D pizza images with \textit{scale}s (range=$[1.0, 3.0]$), x-shift (range=$[-112,112]$ shown as \textit{dx} in figures) and y-shift (range=$[-112,112]$, shown as \textit{dy} in figures). 
We choose this range because the input image size is $224 \times 224$ for the view attribute classifier, so the center of the pizza samples will appear in the image.
\autoref{fig:pizza10_performance_and_pizzaView_examples}(b) illustrates the meaning of the four view attributes.
These transformations are combined together to create arbitrarily many samples for training of the view attribute regressor.

We use ResNext101~\cite{xie2017aggregated} as the backbone, mean square error as the loss term, and batch size of $64$, $75\%$ 3D models are used in training while the rest are in validation. The regressor converges after about $150k$ samples. After training, we use root-mean-square error ($RMSE_{attr}$) as the metric and result is shown in \autoref{tab:quantitative_comparison} (last row), these errors are very small considering the range of the view attributes.

\section{Experiments}
\label{sec:experiments}
\vspace{-0.5em}

\noindent \textbf{Network Structures.} We use the structures in StyleGAN2 and decrease the style noise dimension from $512$ to $256$. 
The proposed \MSMAE is composed of a group of sub-encoders, each sub-encoder is a fully-connected layer with ReLU activation, the label embedding dimension is set to be $T=256$, hence, after concatenating with style noise $w$, the input to the generator is of the same dimension as that in StyleGAN2.

\vspace{0.2em}
\noindent \textbf{Configurations.} For the generator, $\lambda_{ingr} = \lambda_{view}=1.0$, since there are no view attribute labels in \pizzaIngr, we use predicted view attributes as the ground truth label when training the generator. 
For the discriminator, $\lambda_{r1} = 10$, R1 regularization is performed every 16 mini-batches as in StyleGAN2. 
The learning rate for both generator and discriminator is $0.002$. 


\vspace{0.2em}
\noindent \textbf{Metrics.} We use FID~\cite{heusel2017gans} (on 5K images) to assess quality of the generated images; lower FID indicates the generated image distribution is closer to the real image distribution. 
We use the mean Average Precision~\cite{zhu2004recall} ($mAP$) to evaluate the conditioning on the desired ingredients. The ingredient classifier $h_{ingr}$ used is trained on \pizzaIngr as shown in \autoref{sec:pizza10} and achieves $mAP=0.6804$. 
We finally employ root-mean-square error (RMSE) to evaluate the conditioning on the desired view attributes, where the view regressor $h_{view}$ is trained on \pizzaView. 
\vspace{-1em}

\subsection{Comparison with Baselines}
\label{subsec:compare_with_baselines}

\begin{figure}[t]
    \centering
    \begin{minipage}[t]{0.48\textwidth}
        \includegraphics[width=\textwidth]{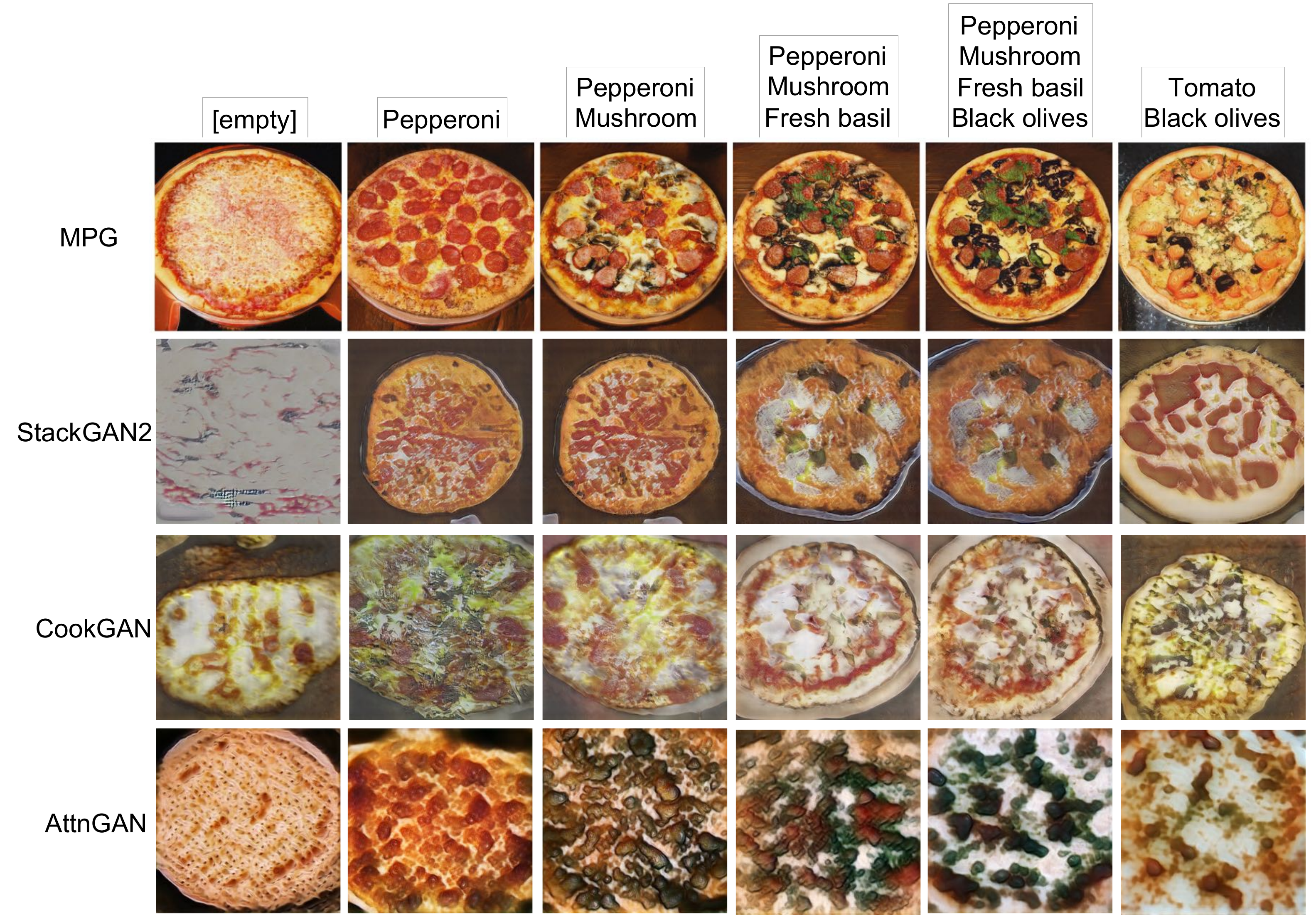}
        \vspace{0.01em}
        \caption*{(a)}
    \end{minipage}
    \hfill
    \begin{minipage}[t]{0.48\textwidth}
        \centering
        \includegraphics[width=\textwidth]{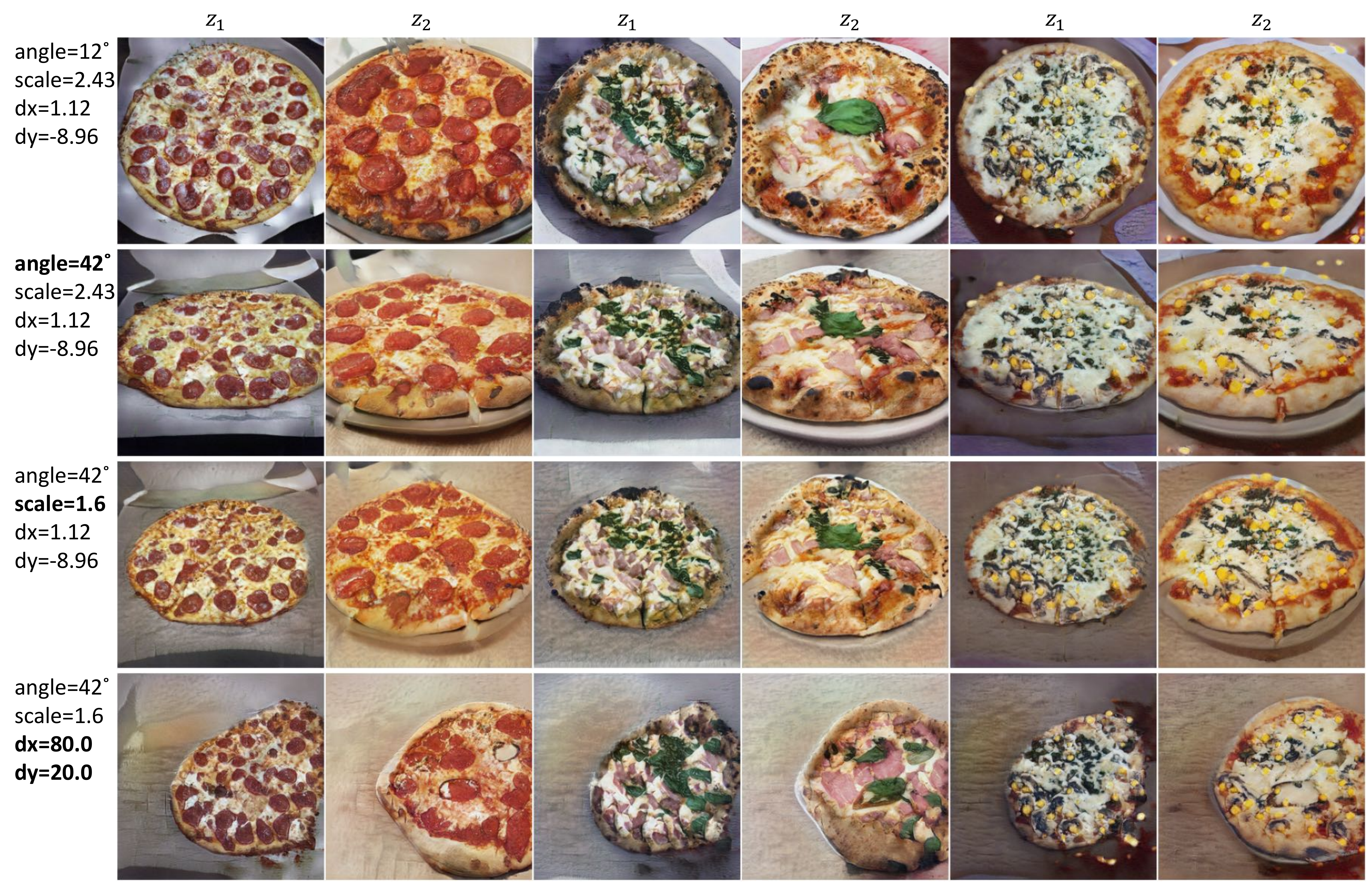}
        \vspace{0.01em}
        \caption*{(b)}
    \end{minipage}
    \vspace{0.5em}
    \caption{(a) Qualitative comparison between \MPG and baselines. (b) Images generated using \MPG from different combinations of ingredients, view attributes and style noise. Images in each row are of the same view attributes and images in each column are of the same style noise. Please read \autoref{subsec:assessment} for more details}
    \vspace{-1em}
    \label{fig:baselines_and_independence}
\end{figure}
\vspace{-1em}

We first compare \MPG with three works on text-based image generation, where we use the concatenated ingredient names as the text description.  
View attributes are left out in this comparison.
\textbf{StackGAN2}~\cite{zhang2018stackgan++} extracts the text feature from a pretrained retrieval model then forwards it through a stacked generator with a separate discriminators working at different scales.
\textbf{CookGAN}~\cite{han2020cookgan} also applies StackGAN2 with a cycle-consistent loss to drive the generated images to retrieve its desired ingredients. 
\textbf{AttnGAN}~\cite{xu2018attngan} uses a pretrained text encoder to capture the importance of each word and leverages both word-level and sentence-level features to guide the generation process.


\autoref{tab:quantitative_comparison} compares between \MPG and the baselines. \MPG offers a huge leap forward compared with prior SoTA both in terms of FID and mAP, suggesting it not only generates realistic images with diversity but also improves the conditioning on the desired ingredients.

FID of \MPG approaches the lower bound of $8.20$, which can be estimated from real images\footnote{Obtained by randomly splitting \pizzaIngr into two subsets of about 5K images each, then computing the FID between the two sets.}. 
Note that the mAP is much higher on the  images generated from \MPG, even when compared with that on real images ($0.6804$). While this may, at first, appear questionable, we attribute it to the model trained with limited data. There, the synthesized images capture the distinct characteristics of each ingredient but are incapable of covering the real image diversity, making the retrieval easier than that with real images.

\begin{table*}[t]
    \centering
    \scriptsize
    \begin{tabular}{ccccccc}
        \toprule
        Model & FID$\downarrow$ & mAP$_{ingr}\uparrow$ & RMSE$_{angle}\downarrow$ & RMSE$_{scale}\downarrow$ & RMSE$_{dx}\downarrow$ & RMSE$_{dy}\downarrow$\\
        \midrule
        StackGAN2~\cite{zhang2018stackgan++} & $63.51$ & $0.3219$ &-&-&-&-\\
        CookGAN~\cite{han2020cookgan} & $45.64$ & $0.2796$ &-&-&-&-\\
        AttnGAN~\cite{xu2018attngan} & $74.46$ & $0.5729$ &-&-&-&-\\
        \MPG  & $\mathbf{7.44}$ & $\mathbf{0.9998}$  & $\mathbf{3.39}$ & $\mathbf{0.09}$ & $\mathbf{4.28}$ & $\mathbf{4.09}$\\
        \MPG-\MSMAE & $8.13$ & $0.9987$ & $3.69$ & $0.16$ & $4.76$ & $4.24$\\
        \MPG-\MSMAE* & $9.37$ & $0.9867$ & $5.32$ & $0.21$ & $6.94$ & $9.12$\\
        \MPG-\AR  & $7.48$ & $0.2404$ & $17.04$ & $0.34$ & $15.54$ & $14.94$\\
        \midrule
        ingr\_cls on \pizzaIngr & - & 0.6804 & - & - & - & - \\
        view\_reg on \pizzaView & - & - & $1.70$ & $0.03$ & $1.81$ & $1.84$ \\
        \bottomrule
    \end{tabular}
    \vspace{1em}
    \caption{Quantitative comparison of performances between baselines, \MPG and its counterparts with missing components}
    \vspace{-2em}
    \label{tab:quantitative_comparison}
\end{table*}


\autoref{fig:baselines_and_independence}(a) shows samples from our \MPG as well as the baselines conditioned on sample ingredient lists. 
\MPG images are subjectively more appealing, with the desired ingredients, compared to the baselines. Notice how \MPG can effectively add \textit{Pepperoni, Mushroom, Fresh basil and Black olives} one after another into the generated images, more examples are provided in the supplementary materials.


\subsection{Ablation Study}
\label{subsec:ablation_study}

To verify the design of \textbf{\MPG}, we consider the following ablation experiments. 
\textbf{\MPG-\MSMAE} removes Multi-Scale Multi-Attribute Encoder, using the same label embedding for all resolutions. 
\textbf{\MPG-\MSMAE*} is similar but the label embedding is concatenated with the style noise $\z$ and forwarded through the mapping network, instead of concatenating it with the output $w$ from the mapping network. 
\textbf{\MPG-\AR} removes the Attribute Regularizer.

From \autoref{tab:quantitative_comparison} we observe that removing \textbf{\MSMAE} increases FID and decreases mAP and RMSE, while removing Attribute Regularizer greatly decreases mAP and RMSE. This is intuitive, with \AR the only component to control the attributes during training. In the supplementary materials, we also compare with SeFa~\cite{shen2020closed}, which tries to find semantically meaningful dimensions that control attributes in StyleGAN by eigen decomposition.

\begin{figure*}[ht]
    \centering
    \includegraphics[width=\textwidth]{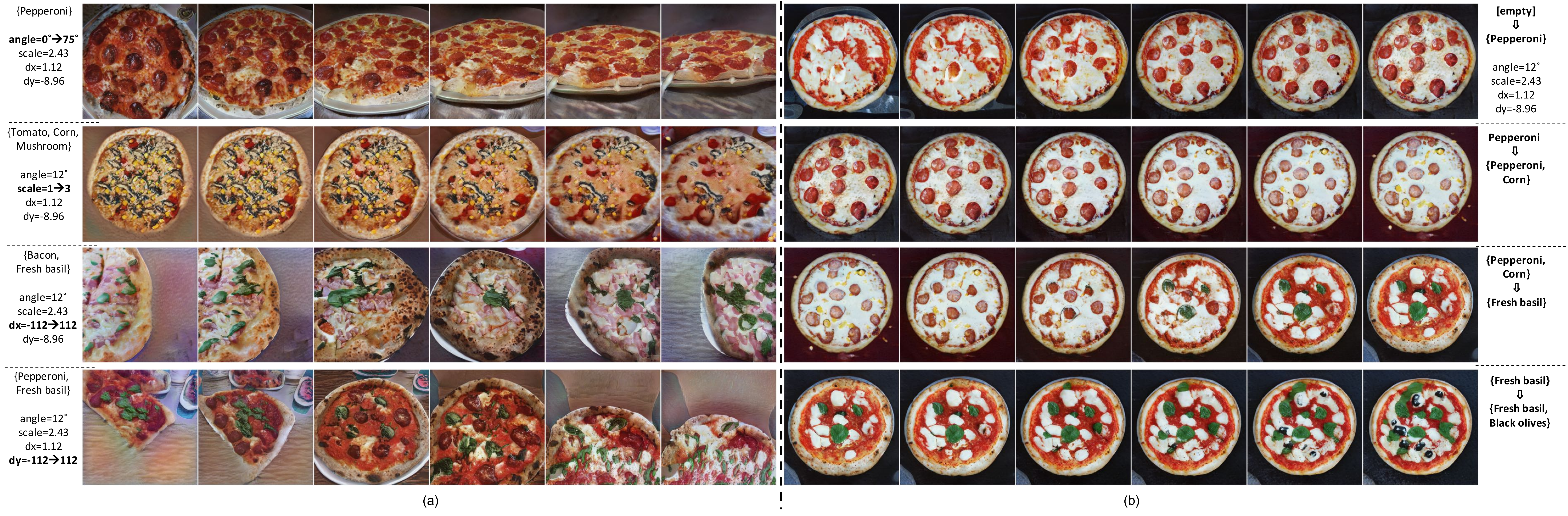}
    \caption{Traversing view attributes (a) and ingredient attributes (b). To traverse the  the view attributes (a), images in each row are generated by changing one view attribute at a time (left-most legend column) while fixing the ingredients and the style noise. To traverse the ingredient attributes (b), images in each row are generated by adding or removing select ingredients (right-most column) while keeping the view attributes and the style noise fixed.}
    \vspace{-1em}
    \label{fig:smoothness}
\end{figure*}
\vspace{-1em}

\subsection{Assessment of Generated Images}
\label{subsec:assessment}
We provide additional analyses of images created by \MPG. We first qualitatively study the independence between label embedding and the style noise, then assess the smoothness of the text and the style noise spaces. Additional experiment in the supplementary materials accesses generated image quality with uncommon view attributes more thoroughly.

\vspace{0.2em}
\noindent \textbf{Independence.} 
To analyze the independence between input components, we generate images by fixing each component (i.e., ingredients, view attributes and style noise $z$). 
The result is shown in \autoref{fig:baselines_and_independence}(b); images in each row share the same view attributes while each column corresponds to the same style noise $z$. Pairs of adjacent columns are generated with identical ingredients (i.e., \{pepperoni\}, \{bacon, fresh basil\}, \{mushroom, corn\}).

The key observations are: (1) Across each column, view attributes can be modified effectively according to the view attribute input. (2) In each row, images tend to exhibit similar view attributes, differing in color and shape of the pizza and background. These demonstrate that view attribute input and style noise $z$ are independent of each other. (3) By comparing images generated with identical view attributes and style noise $z$ but different ingredients (\eg [row 1, column 1] vs. [row1, column 3] vs [row1, column 5]), we  observe that pizzas have similar view attributes, shape and color, the backgrounds are also alike, which demonstrates the independence of the ingredient input and the view attributes and style noise $z$.

\vspace{0.2em}
\noindent \textbf{Smoothness.} 
We verify the smoothness of the learned embedding space by traversing the input space. 
In \autoref{fig:smoothness} (a), along each row, we fix the ingredients and style noise, while setting the three view attributes to their most frequently predicted values and traverse along one view attribute. It can be seen the targeting view attribute changes smoothly in each row.
Note that large $dx$ and $dy$ could result in the change of the image style (\eg pizza shape); we believe this is caused by \pizzaIngr, used to train \MPG, not containing sufficiently many pizza images with similar view attributes. This additionally explains why it is hard to generate images as small as shown in \autoref{fig:pizza10_performance_and_pizzaView_examples}(b) left.
In \autoref{fig:smoothness} (b), along each row, we fix the view attributes and style noise and traverse along the ingredient input; the traversing is conducted by treating the binary indicator existence index as a floating-point number(range=[0.0, 1.0]) and interpolating in between.
Each row follows the previous by gradually adding or removing certain ingredients. This suggests the ingredients also change smoothly.
Together, this demonstrates the ability of \MPG to learn, then recreate independent and smooth major factors of variation in this complex image space.
\vspace{-2em}
\
\section{Conclusion}
\label{sec:conclusion}
\vspace{-0.5em}

We propose \MPG, a Multi-attribute Pizza Generator image synthesis framework, supported by two new datasets \pizzaIngr and \pizzaView, that can effectively learn to create photo-realistic pizza images from combinations of explicit ingredients and geometric view attributes. While effective in visual appearance, the model remains computationally expensive to train, with some loss of detail at fine resolutions, the problems we aim to investigated in future research.

\bibliography{egbib}

\begin{thebibliography}{33}
\providecommand{\natexlab}[1]{#1}
\providecommand{\url}[1]{\texttt{#1}}
\expandafter\ifx\csname urlstyle\endcsname\relax
  \providecommand{\doi}[1]{doi: #1}\else
  \providecommand{\doi}{doi: \begingroup \urlstyle{rm}\Url}\fi

\bibitem[Brock et~al.(2018)Brock, Donahue, and Simonyan]{brock2018large}
Andrew Brock, Jeff Donahue, and Karen Simonyan.
\newblock Large scale {GAN} training for high fidelity natural image synthesis.
\newblock \emph{arXiv preprint arXiv:1809.11096}, 2018.

\bibitem[Choi et~al.(2018)Choi, Choi, Kim, Ha, Kim, and Choo]{choi2018stargan}
Yunjey Choi, Minje Choi, Munyoung Kim, Jung-Woo Ha, Sunghun Kim, and Jaegul
  Choo.
\newblock Star{GAN}: Unified generative adversarial networks for multi-domain
  image-to-image translation.
\newblock In \emph{Proceedings of the IEEE conference on computer vision and
  pattern recognition}, pages 8789--8797, 2018.

\bibitem[Drucker and Le~Cun(1992)]{drucker1992improving}
Harris Drucker and Yann Le~Cun.
\newblock Improving generalization performance using double backpropagation.
\newblock \emph{IEEE Transactions on Neural Networks}, 3\penalty0 (6):\penalty0
  991--997, 1992.

\bibitem[El et~al.(2019)El, Licht, and Yosephian]{el2019gilt}
Ori~Bar El, Ori Licht, and Netanel Yosephian.
\newblock Gilt: Generating images from long text.
\newblock \emph{arXiv preprint arXiv:1901.02404}, 2019.

\bibitem[Goodfellow et~al.(2014)Goodfellow, Pouget-Abadie, Mirza, Xu,
  Warde-Farley, Ozair, Courville, and Bengio]{goodfellow2014generative}
Ian Goodfellow, Jean Pouget-Abadie, Mehdi Mirza, Bing Xu, David Warde-Farley,
  Sherjil Ozair, Aaron Courville, and Yoshua Bengio.
\newblock Generative adversarial nets.
\newblock In \emph{Advances in neural information processing systems}, pages
  2672--2680, 2014.

\bibitem[Han et~al.(2020)Han, Guerrero, and Pavlovic]{han2020cookgan}
Fangda Han, Ricardo Guerrero, and Vladimir Pavlovic.
\newblock Cook{GAN}: Meal image synthesis from ingredients.
\newblock In \emph{The IEEE Winter Conference on Applications of Computer
  Vision}, pages 1450--1458, 2020.

\bibitem[He et~al.(2016)He, Zhang, Ren, and Sun]{he2016deep}
Kaiming He, Xiangyu Zhang, Shaoqing Ren, and Jian Sun.
\newblock Deep residual learning for image recognition.
\newblock In \emph{Proceedings of the IEEE conference on computer vision and
  pattern recognition}, pages 770--778, 2016.

\bibitem[He et~al.(2019)He, Zuo, Kan, Shan, and Chen]{he2019attgan}
Zhenliang He, Wangmeng Zuo, Meina Kan, Shiguang Shan, and Xilin Chen.
\newblock Att{GAN}: Facial attribute editing by only changing what you want.
\newblock \emph{IEEE Transactions on Image Processing}, 28\penalty0
  (11):\penalty0 5464--5478, 2019.

\bibitem[Heusel et~al.(2017)Heusel, Ramsauer, Unterthiner, Nessler, and
  Hochreiter]{heusel2017gans}
Martin Heusel, Hubert Ramsauer, Thomas Unterthiner, Bernhard Nessler, and Sepp
  Hochreiter.
\newblock {GAN}s trained by a two time-scale update rule converge to a local
  nash equilibrium.
\newblock In \emph{Advances in neural information processing systems}, pages
  6626--6637, 2017.

\bibitem[Isola et~al.(2017)Isola, Zhu, Zhou, and Efros]{isola2017image}
Phillip Isola, Jun-Yan Zhu, Tinghui Zhou, and Alexei~A Efros.
\newblock Image-to-image translation with conditional adversarial networks.
\newblock In \emph{Proceedings of the IEEE conference on computer vision and
  pattern recognition}, pages 1125--1134, 2017.

\bibitem[Ito et~al.(2018)Ito, Shimoda, and Yanai]{ito2018food}
Yoshifumi Ito, Wataru Shimoda, and Keiji Yanai.
\newblock Food image generation using a large amount of food images with
  conditional {GAN}: ramen{GAN} and recipe{GAN}.
\newblock In \emph{Proceedings of the Joint Workshop on Multimedia for Cooking
  and Eating Activities and Multimedia Assisted Dietary Management}, pages
  71--74, 2018.

\bibitem[Karras et~al.(2017)Karras, Aila, Laine, and
  Lehtinen]{karras2017progressive}
Tero Karras, Timo Aila, Samuli Laine, and Jaakko Lehtinen.
\newblock Progressive growing of {GAN}s for improved quality, stability, and
  variation.
\newblock \emph{arXiv preprint arXiv:1710.10196}, 2017.

\bibitem[Karras et~al.(2019)Karras, Laine, and Aila]{karras2019style}
Tero Karras, Samuli Laine, and Timo Aila.
\newblock A style-based generator architecture for generative adversarial
  networks.
\newblock In \emph{Proceedings of the IEEE conference on computer vision and
  pattern recognition}, pages 4401--4410, 2019.

\bibitem[Karras et~al.(2020{\natexlab{a}})Karras, Aittala, Hellsten, Laine,
  Lehtinen, and Aila]{karras2020training}
Tero Karras, Miika Aittala, Janne Hellsten, Samuli Laine, Jaakko Lehtinen, and
  Timo Aila.
\newblock Training generative adversarial networks with limited data.
\newblock \emph{Advances in Neural Information Processing Systems}, 33,
  2020{\natexlab{a}}.

\bibitem[Karras et~al.(2020{\natexlab{b}})Karras, Laine, Aittala, Hellsten,
  Lehtinen, and Aila]{karras2020analyzing}
Tero Karras, Samuli Laine, Miika Aittala, Janne Hellsten, Jaakko Lehtinen, and
  Timo Aila.
\newblock Analyzing and improving the image quality of {StyleGAN}.
\newblock In \emph{Proceedings of the IEEE/CVF Conference on Computer Vision
  and Pattern Recognition}, pages 8110--8119, 2020{\natexlab{b}}.

\bibitem[Li et~al.(2019)Li, Qi, Lukasiewicz, and Torr]{li2019controllable}
Bowen Li, Xiaojuan Qi, Thomas Lukasiewicz, and Philip Torr.
\newblock Controllable text-to-image generation.
\newblock In \emph{Advances in Neural Information Processing Systems}, pages
  2065--2075, 2019.

\bibitem[Mirza and Osindero(2014)]{mirza2014conditional}
Mehdi Mirza and Simon Osindero.
\newblock Conditional generative adversarial nets.
\newblock \emph{arXiv preprint arXiv:1411.1784}, 2014.

\bibitem[Miyato and Koyama(2018)]{miyato2018cgans}
Takeru Miyato and Masanori Koyama.
\newblock c{GAN}s with projection discriminator.
\newblock \emph{arXiv preprint arXiv:1802.05637}, 2018.

\bibitem[Miyato et~al.(2018)Miyato, Kataoka, Koyama, and
  Yoshida]{miyato2018spectral}
Takeru Miyato, Toshiki Kataoka, Masanori Koyama, and Yuichi Yoshida.
\newblock Spectral normalization for generative adversarial networks.
\newblock \emph{arXiv preprint arXiv:1802.05957}, 2018.

\bibitem[Odena et~al.(2017)Odena, Olah, and Shlens]{odena2017conditional}
Augustus Odena, Christopher Olah, and Jonathon Shlens.
\newblock Conditional image synthesis with auxiliary classifier {GAN}s.
\newblock In \emph{International conference on machine learning}, pages
  2642--2651, 2017.

\bibitem[Oeldorf and Spanakis(2019)]{oeldorf2019loganv2}
Cedric Oeldorf and Gerasimos Spanakis.
\newblock {LoGANv2: Conditional Style-Based Logo Generation with Generative
  Adversarial Networks}.
\newblock In \emph{2019 18th IEEE International Conference On Machine Learning
  And Applications (ICMLA)}, pages 462--468. IEEE, 2019.

\bibitem[Papadopoulos et~al.(2019)Papadopoulos, Tamaazousti, Ofli, Weber, and
  Torralba]{papadopoulos19cvpr}
Dim~P. Papadopoulos, Youssef Tamaazousti, Ferda Ofli, Ingmar Weber, and Antonio
  Torralba.
\newblock How to make a pizza: Learning a compositional layer-based {GAN}
  model.
\newblock In \emph{The IEEE Conference on Computer Vision and Pattern
  Recognition (CVPR)}, June 2019.

\bibitem[Qiao et~al.(2019)Qiao, Zhang, Xu, and Tao]{qiao2019mirrorgan}
Tingting Qiao, Jing Zhang, Duanqing Xu, and Dacheng Tao.
\newblock Mirror{GAN}: Learning text-to-image generation by redescription.
\newblock In \emph{Proceedings of the IEEE Conference on Computer Vision and
  Pattern Recognition}, pages 1505--1514, 2019.

\bibitem[Sabini et~al.()Sabini, Abdullah, and Phan]{sabinigan}
Mark Sabini, Zahra Abdullah, and Darrith Phan.
\newblock {GAN}-stronomy: Generative cooking with conditional {DCGAN}s.

\bibitem[Shen and Zhou(2020)]{shen2020closed}
Yujun Shen and Bolei Zhou.
\newblock Closed-form factorization of latent semantics in {GAN}s.
\newblock \emph{arXiv preprint arXiv:2007.06600}, 2020.

\bibitem[Shoshan et~al.(2021)Shoshan, Bhonker, Kviatkovsky, and
  Medioni]{shoshan2021gancontrol}
Alon Shoshan, Nadav Bhonker, Igor Kviatkovsky, and Gerard Medioni.
\newblock {GAN-Control: Explicitly Controllable GANs}, 2021.

\bibitem[Wang et~al.(2019)Wang, Gao, Zhu, Zhang, and Chen]{wang2019food}
Su~Wang, Honghao Gao, Yonghua Zhu, Weilin Zhang, and Yihai Chen.
\newblock A food dish image generation framework based on progressive growing
  {GAN}s.
\newblock In \emph{International Conference on Collaborative Computing:
  Networking, Applications and Worksharing}, pages 323--333. Springer, 2019.

\bibitem[Xie et~al.(2017)Xie, Girshick, Doll{\'a}r, Tu, and
  He]{xie2017aggregated}
Saining Xie, Ross Girshick, Piotr Doll{\'a}r, Zhuowen Tu, and Kaiming He.
\newblock Aggregated residual transformations for deep neural networks.
\newblock In \emph{Proceedings of the IEEE conference on computer vision and
  pattern recognition}, pages 1492--1500, 2017.

\bibitem[Xu et~al.(2018)Xu, Zhang, Huang, Zhang, Gan, Huang, and
  He]{xu2018attngan}
Tao Xu, Pengchuan Zhang, Qiuyuan Huang, Han Zhang, Zhe Gan, Xiaolei Huang, and
  Xiaodong He.
\newblock Attn{GAN}: Fine-grained text to image generation with attentional
  generative adversarial networks.
\newblock In \emph{Proceedings of the IEEE conference on computer vision and
  pattern recognition}, pages 1316--1324, 2018.

\bibitem[Zhang et~al.(2018)Zhang, Xu, Li, Zhang, Wang, Huang, and
  Metaxas]{zhang2018stackgan++}
Han Zhang, Tao Xu, Hongsheng Li, Shaoting Zhang, Xiaogang Wang, Xiaolei Huang,
  and Dimitris~N Metaxas.
\newblock Stack{GAN}++: Realistic image synthesis with stacked generative
  adversarial networks.
\newblock \emph{IEEE transactions on pattern analysis and machine
  intelligence}, 41\penalty0 (8):\penalty0 1947--1962, 2018.

\bibitem[Zhu and Ngo(2020)]{zhu2020cookgan}
Bin Zhu and Chong-Wah Ngo.
\newblock {CookGAN: Causality based Text-to-Image Synthesis}.
\newblock In \emph{Proceedings of the IEEE/CVF Conference on Computer Vision
  and Pattern Recognition}, pages 5519--5527, 2020.

\bibitem[Zhu et~al.(2017)Zhu, Park, Isola, and Efros]{zhu2017unpaired}
Jun-Yan Zhu, Taesung Park, Phillip Isola, and Alexei~A Efros.
\newblock Unpaired image-to-image translation using cycle-consistent
  adversarial networks.
\newblock In \emph{Proceedings of the IEEE international conference on computer
  vision}, pages 2223--2232, 2017.

\bibitem[Zhu(2004)]{zhu2004recall}
Mu~Zhu.
\newblock Recall, precision and average precision.
\newblock \emph{Department of Statistics and Actuarial Science, University of
  Waterloo, Waterloo}, 2:\penalty0 30, 2004.

\end{thebibliography}

\clearpage
\newpage
\appendix
\section{Visual Comparisons with Baselines}
\begin{figure*}
    \centering
    \begin{minipage}{0.475\textwidth}
        \centering
        \includegraphics[width=\textwidth]{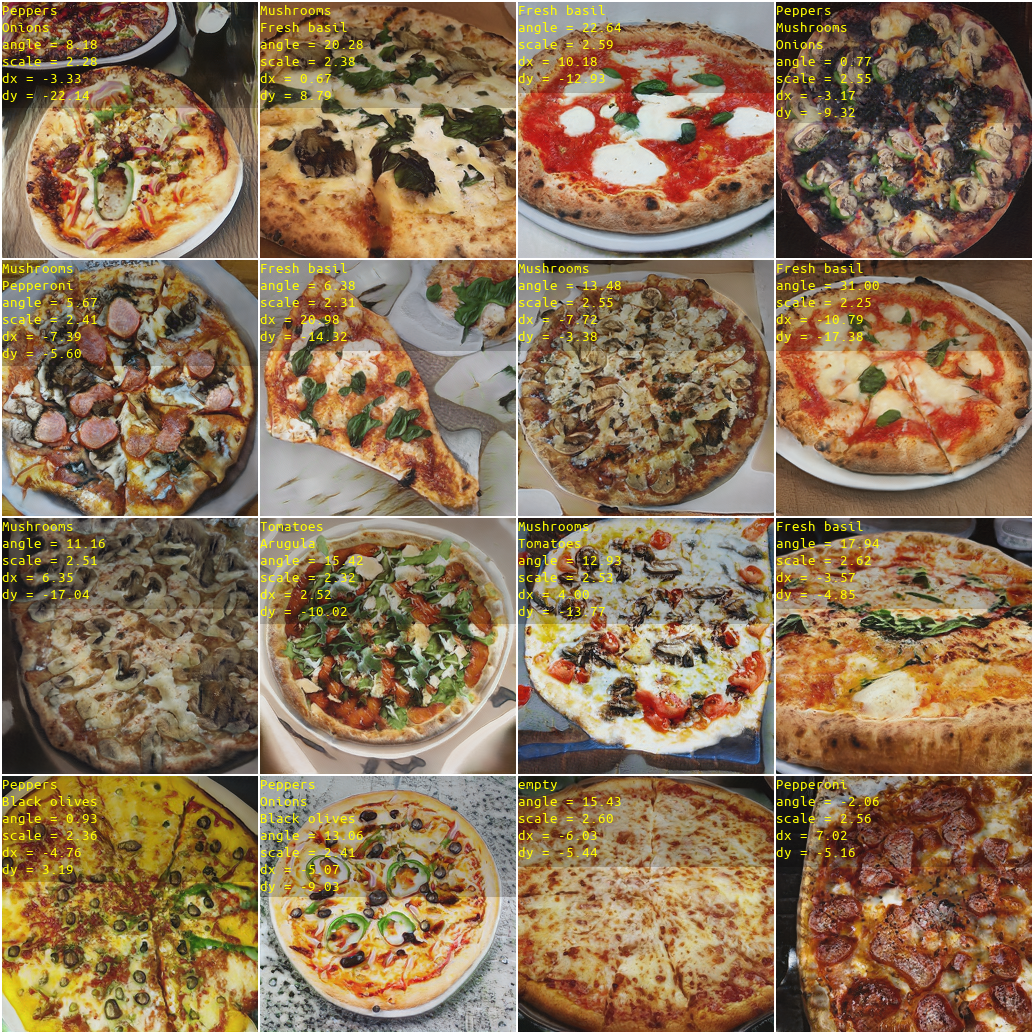}
        \vspace{0.2em}
        \caption*{\small(a) \MPG}
    \end{minipage}
    \hfill
    \begin{minipage}{0.475\textwidth}  
        \centering 
        \includegraphics[width=\textwidth]{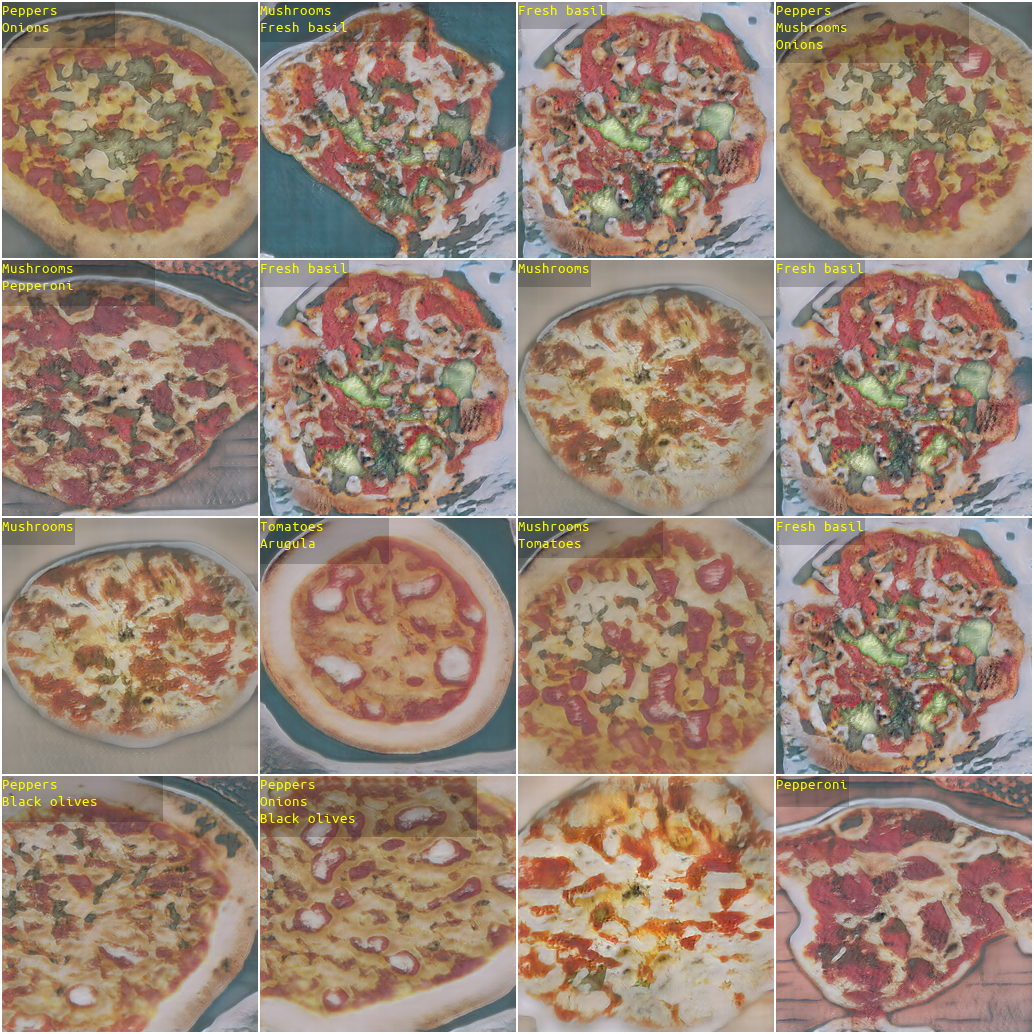}
        \vspace{0.2em}
        \caption*{\small(b) StackGAN2~\cite{zhang2018stackgan++}}
    \end{minipage}
    \vskip\baselineskip
    \begin{minipage}{0.475\textwidth}   
        \centering 
        \includegraphics[width=\textwidth]{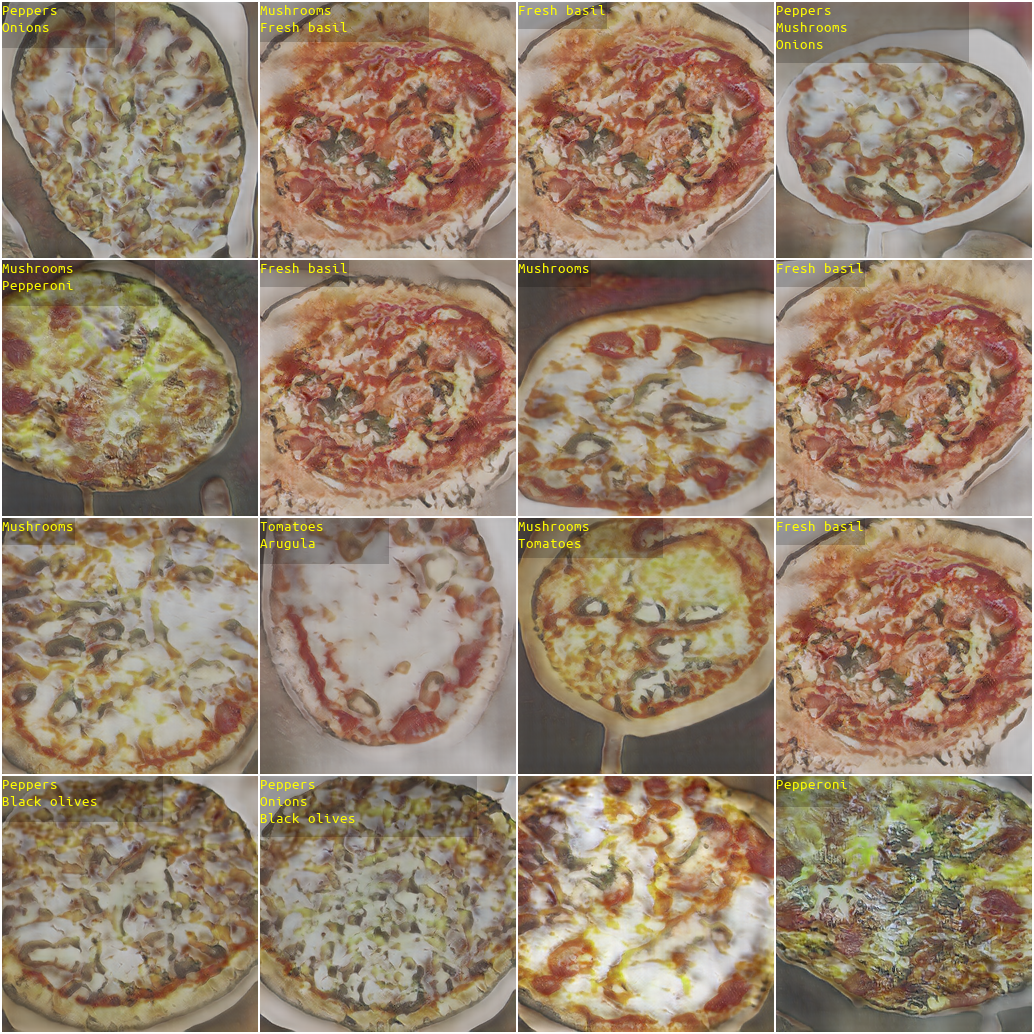}
        \vspace{0.2em}
        \caption*{\small(c) CookGAN~\cite{han2020cookgan}}
    \end{minipage}
    \hfill
    \begin{minipage}{0.475\textwidth}   
        \centering 
        \includegraphics[width=\textwidth]{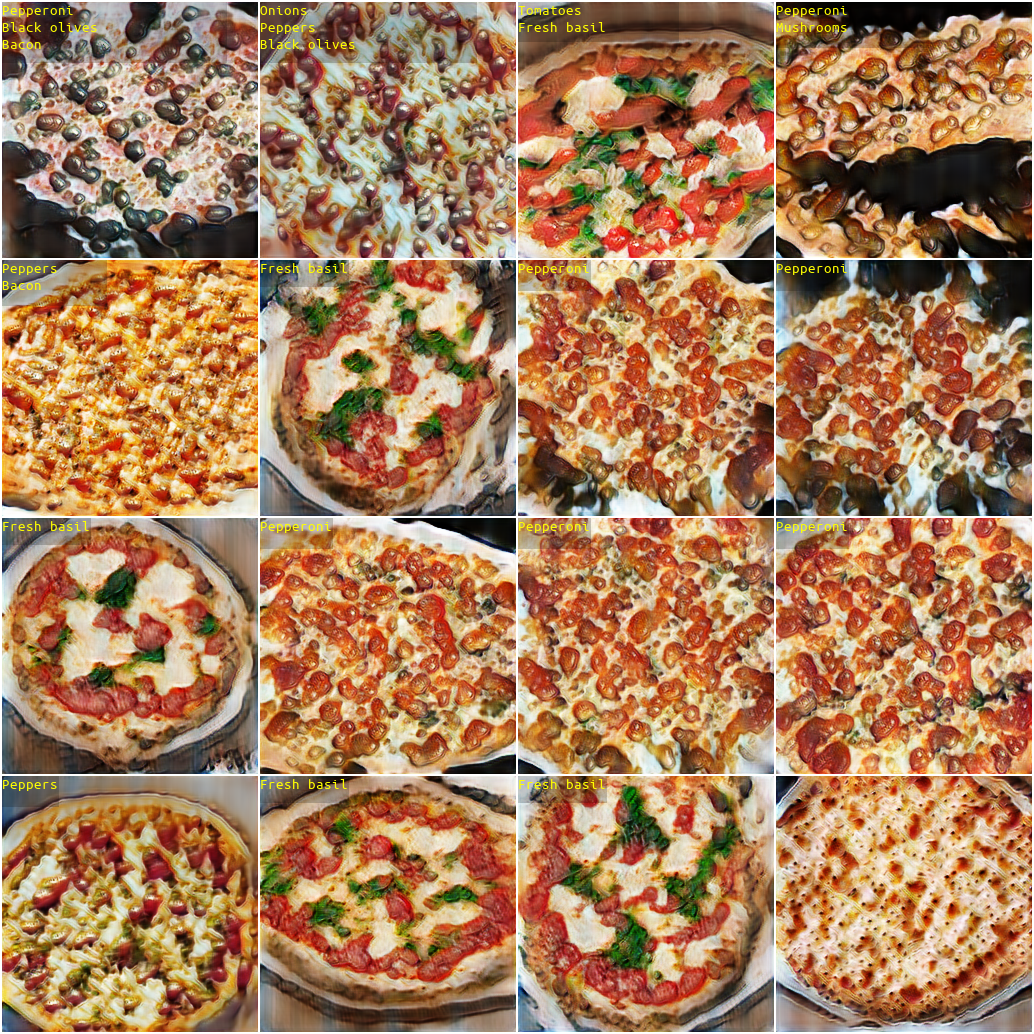}
        \vspace{0.2em}
        \caption*{\small(d) AttnGAN~\cite{xu2018attngan}}
    \end{minipage}
    \vspace{1em}
    \caption{Visual comparisons with baselines. View attributes and style noise are randomly sampled, with the ingredients and view attribute values shown in the top left corner of each image. Note that images generated by \MPG have better quality than the baselines, with the desired ingredients more prominently displayed, without the visual distortions present in the competing approaches.} 
    \label{fig:comparison_with_baselines}
\end{figure*}

We provide additional samples from the proposed \MPG and other state-of-the-art models to further prove the effectiveness of our model:

\begin{itemize}
    \item \autoref{fig:comparison_with_baselines}(a): Example images generated using our \MPG.
    \item \autoref{fig:comparison_with_baselines}(b): Example images generated using StackGAN2~\cite{zhang2018stackgan++}.
    \item \autoref{fig:comparison_with_baselines}(c): Example images generated using CookGAN~\cite{han2020cookgan}.
    \item \autoref{fig:comparison_with_baselines}(d): Example images generated using AttnGAN~\cite{xu2018attngan}.
\end{itemize}

\section{Traversing Attributes}
As shown in Fig.1 of our main paper, the input of \MPG is a triplet: food content (ingredients), geometric style (view attributes including view point, scale, horizontal and vertical shift), and visual style (diversity in fine-grained visual appearance of ingredients and the final dish).
Here we provide examples of fixing one attribute group, and traversing through the other two groups. These examples provide additional qualitative evidence of the independence between the attribute groups and the smoothness of feature space learned by \MPG.

\begin{figure*}
    \centering
    \begin{minipage}[b]{0.475\textwidth}
        \centering
        \includegraphics[width=\textwidth]{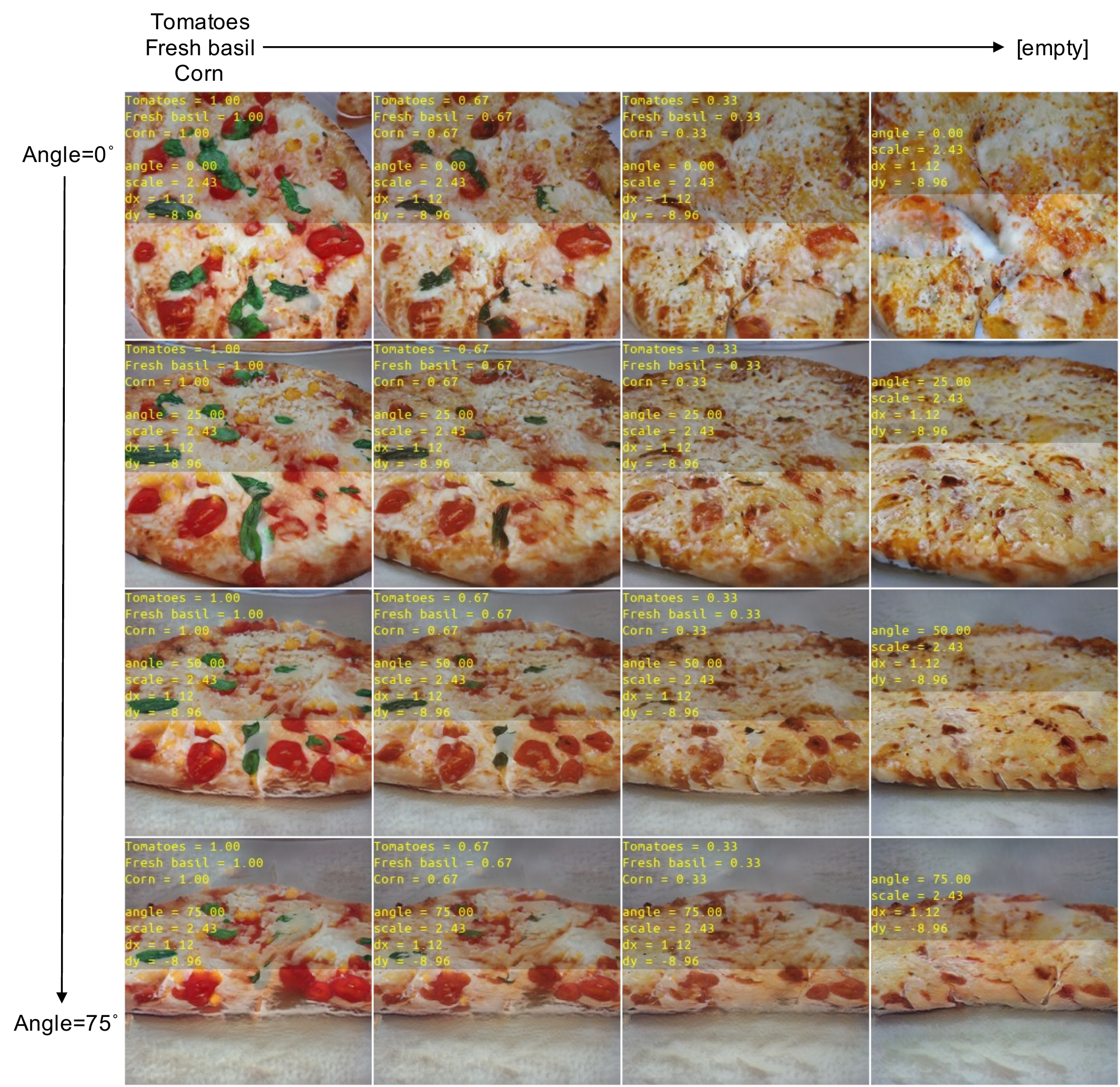}
        \vspace{0.2em}
        \caption*%
        {(a) ingredients ($\rightarrow$) and angle ($\downarrow$)}
        \label{fig:ingr+view_angle}
    \end{minipage}
    \hfill
    \begin{minipage}[b]{0.475\textwidth}  
        \centering 
        \includegraphics[width=\textwidth]{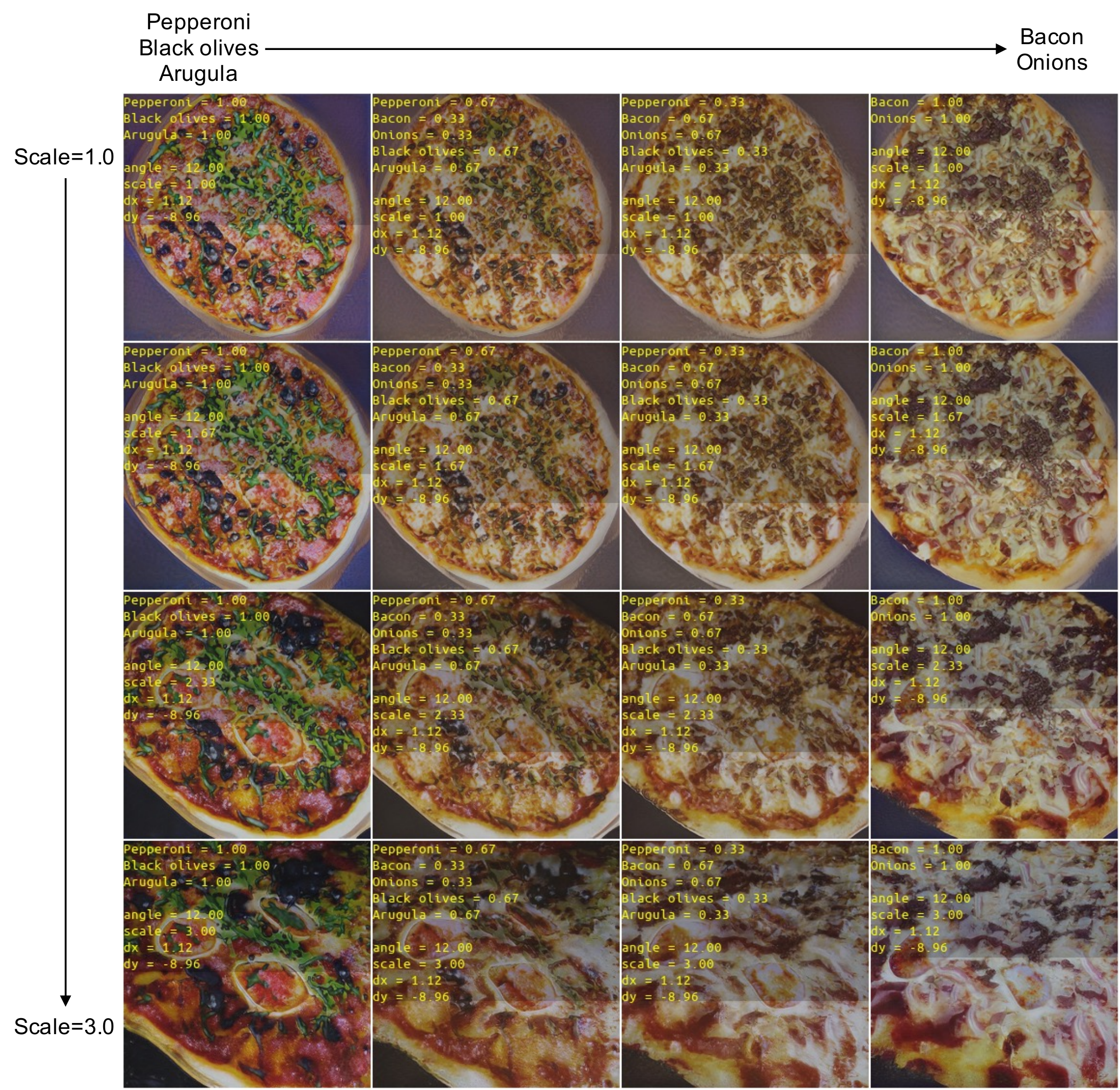}
        \vspace{0.2em}
        \caption*%
        {(b) ingredients ($\rightarrow$) and scale ($\downarrow$)}
        \label{fig:ingr+scale}
    \end{minipage}
    \vskip\baselineskip
    \begin{minipage}[b]{0.475\textwidth}   
        \centering 
        \includegraphics[width=\textwidth]{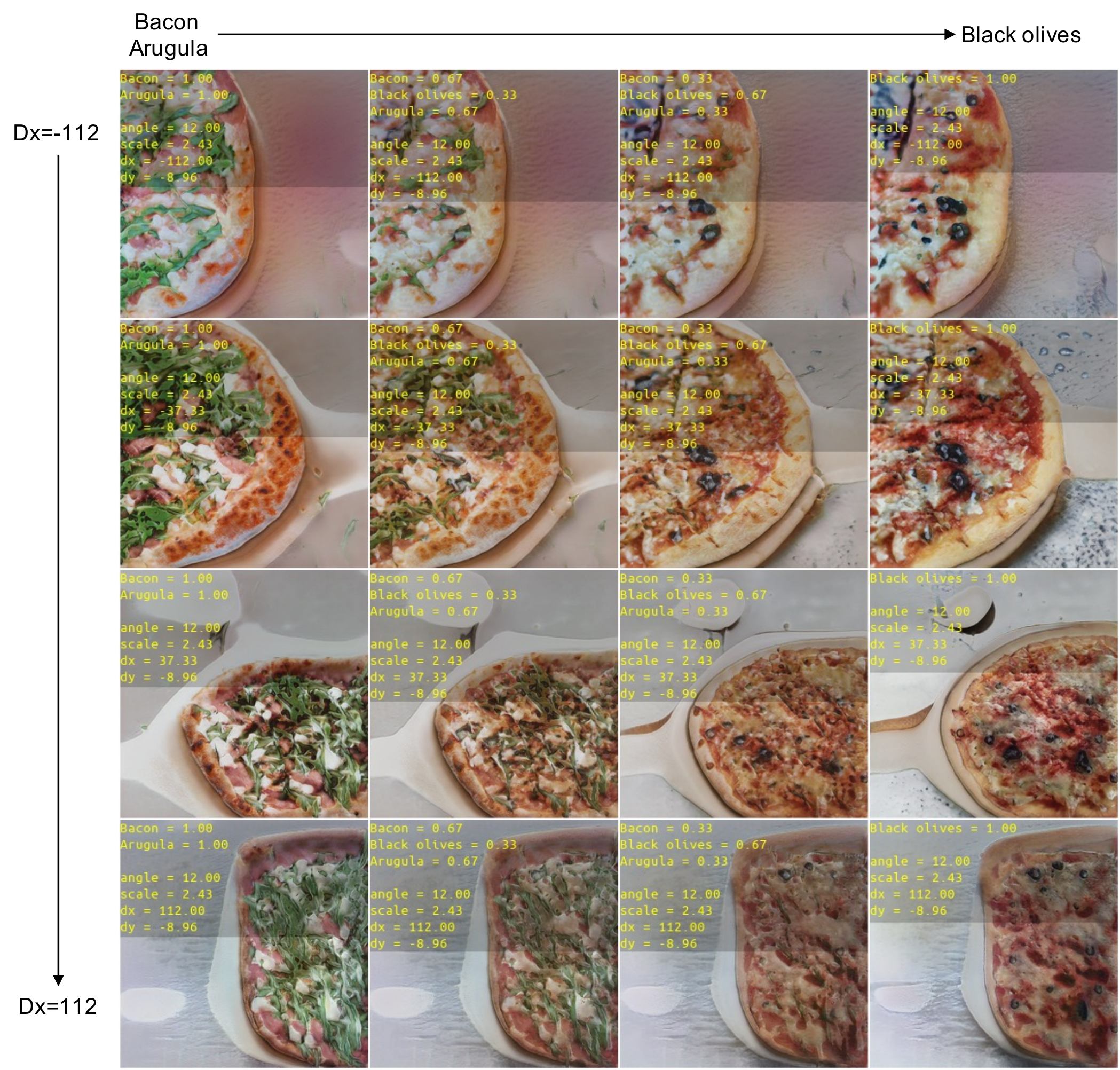}
        \vspace{0.2em}
        \caption*%
        {(c) ingredients ($\rightarrow$) and dx ($\downarrow$)}
        \label{fig:ingr+view_dx}
    \end{minipage}
    \hfill
    \begin{minipage}[b]{0.475\textwidth}   
        \centering 
        \includegraphics[width=\textwidth]{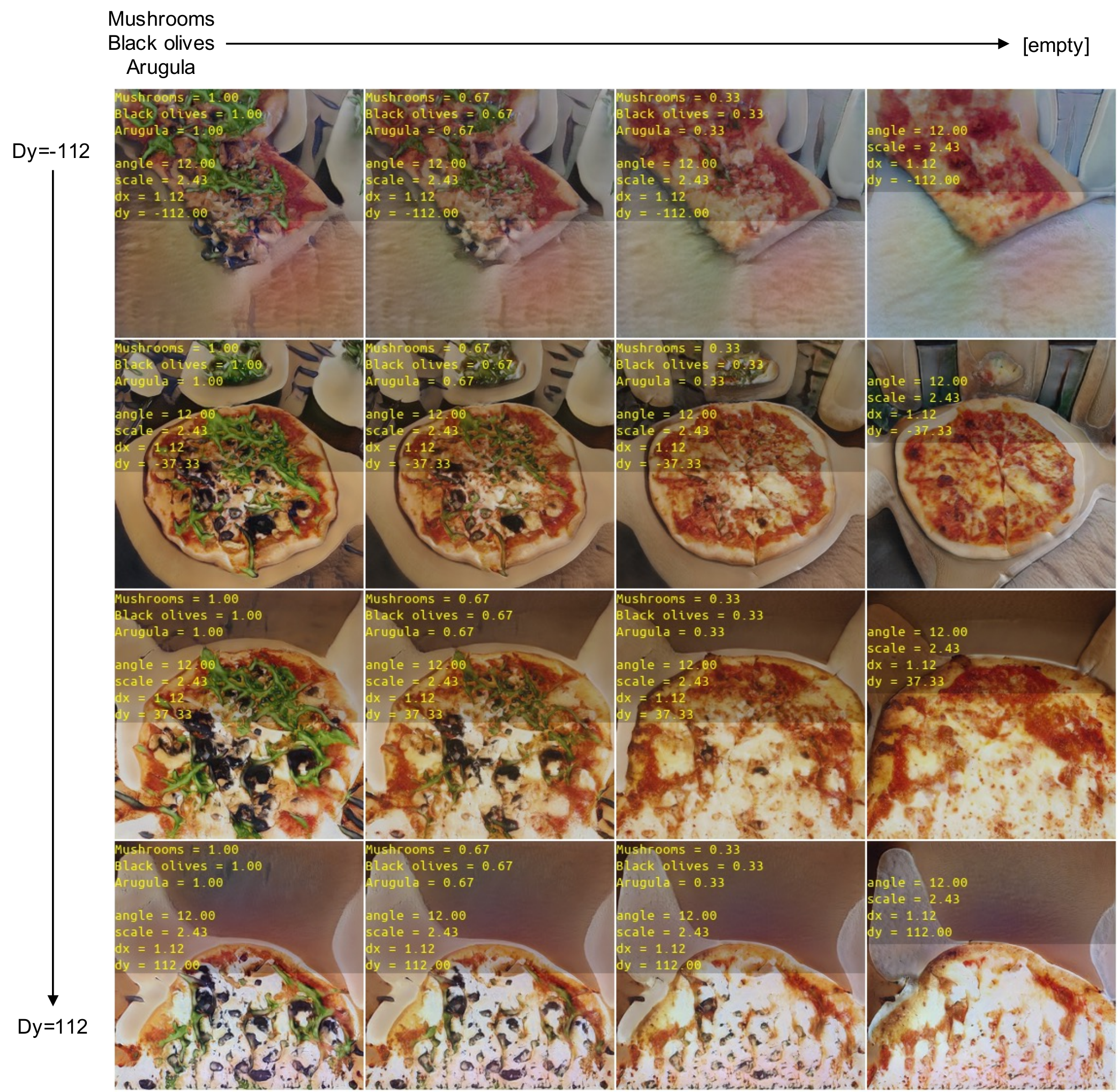}
        \vspace{0.2em}
        \caption*
        {(d) ingredients ($\rightarrow$) and dy ($\downarrow$)}
        \label{fig:ingr+view_dy}
    \end{minipage}
    \vspace{1em}
    \caption[]
    {Images generated using \MPG by fixing the visual style and traversing through \textbf{ingredients} (along horizontal axis) and a \textbf{single view attribute} (along vertical axis, with the other three view attributes fixed). Observe that in each sub-figure, pizza coloring, shapes, and image backgrounds are consistent in response to the desired, fixed visual style. The ingredients and the view attributes smoothly change as we gradually adjust the changing attributes from one end of the range to the other.  For instance, in (a) the viewing angle of the pizza varies smoothly from top to bottom, but each column has the same toppings, starting from tomatoes and basil on the left to the pizza with no toppings on the right.}
    \label{fig:traversing_attributes}
\end{figure*}

\begin{figure}[ht]
    \centering
    \includegraphics[width=0.48\textwidth]{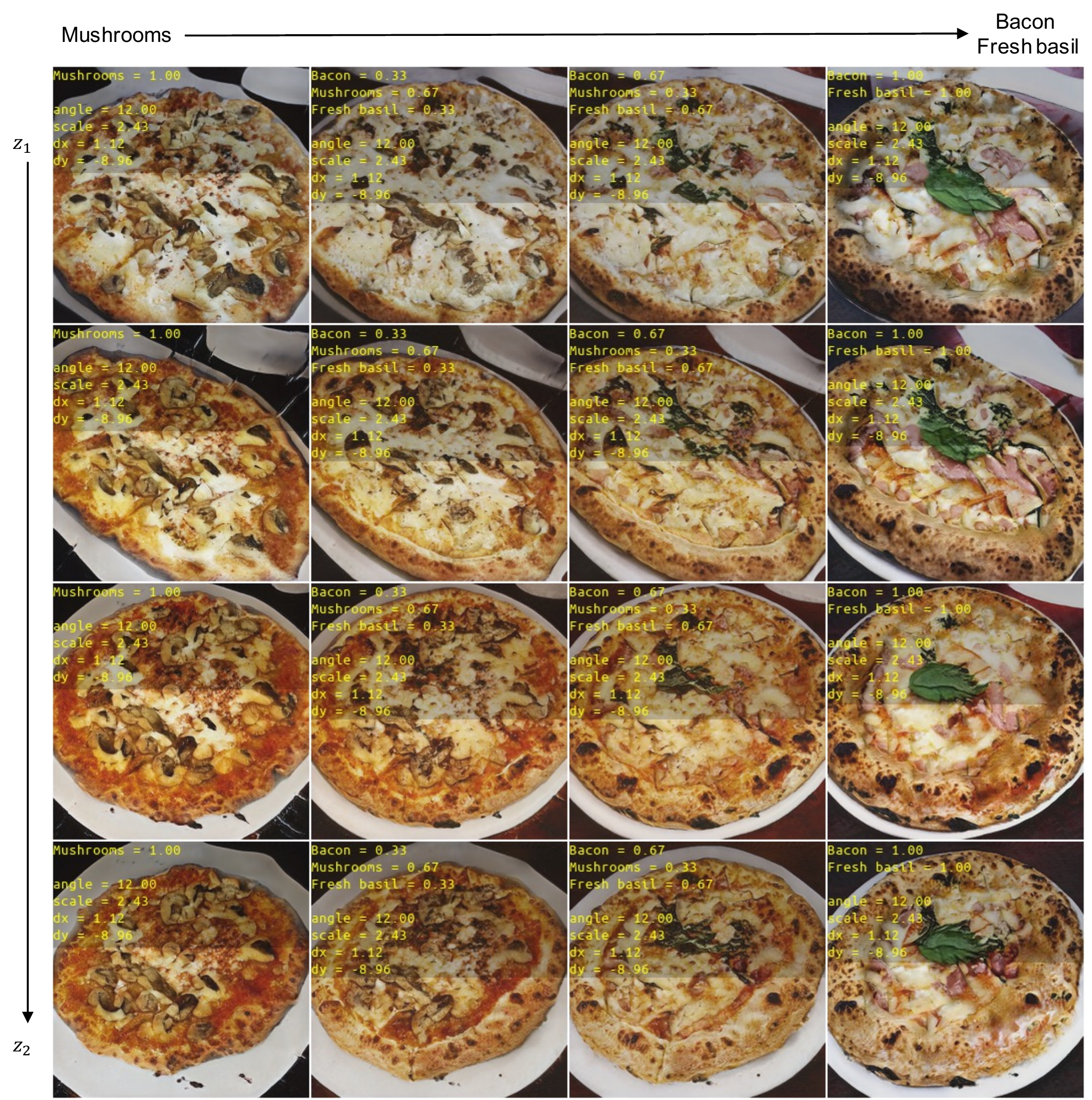}
    \vspace{1em}
    \caption{Images generated using our \MPG by fixing the view attributes and traversing the \textbf{ingredients} (along horizontal axis) and the \textbf{visual style} (along vertical axis). Note that the visual style changing from $z_1$ to $z_2$ leads to a rotated, reddish pizza image turning elliptical for some $z$, while our four view attributes (view point, scale, shift) remain the same. The consistency is retained when we change the ingredients, across columns.}
    \label{fig:ingr+z}
\end{figure}

\begin{figure*}
        \centering
        \begin{minipage}[b]{0.475\textwidth}
            \centering
            \includegraphics[width=\textwidth]{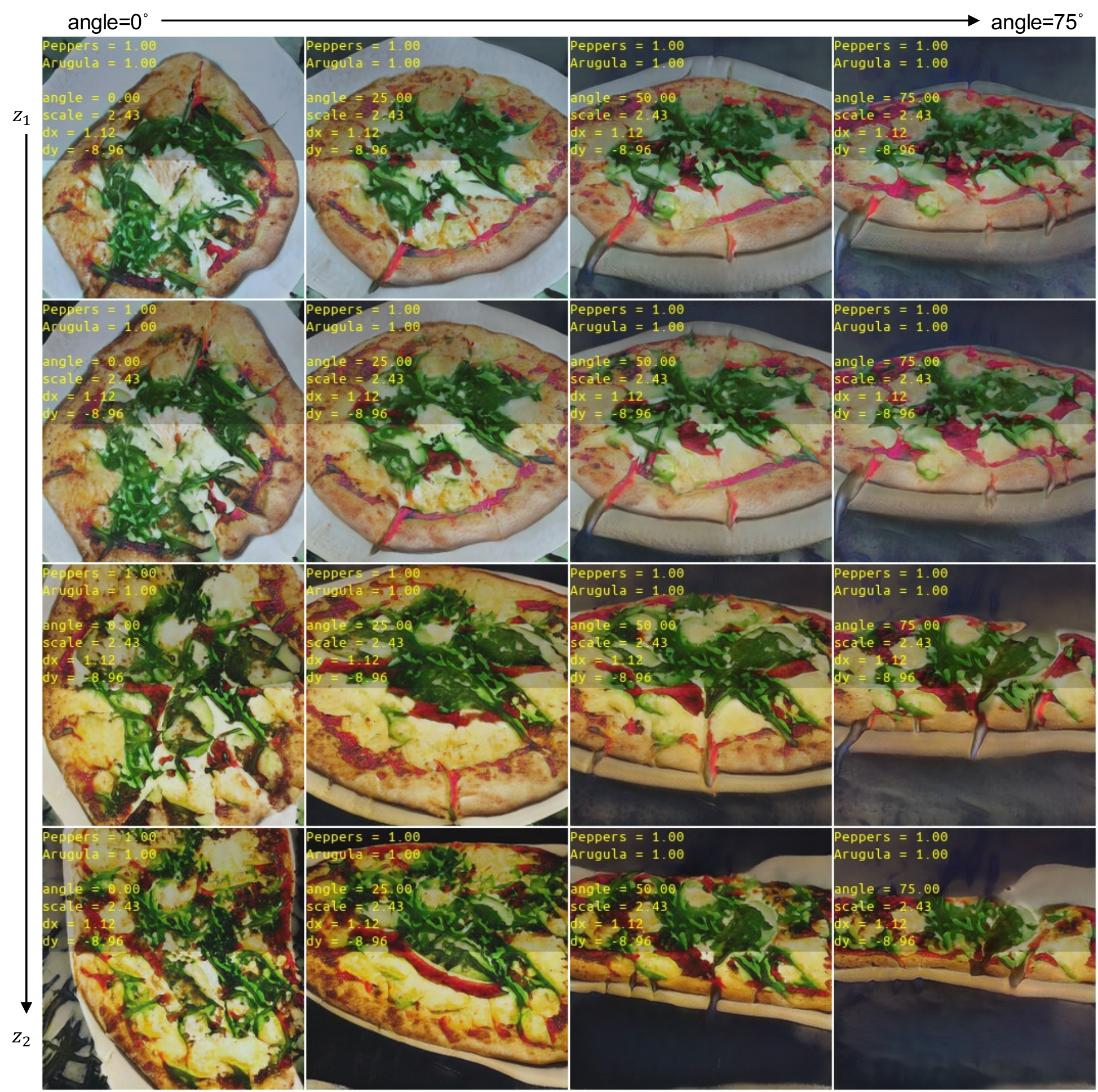}
            \vspace{0.2em}
            \caption*%
            {(a) angle ($\rightarrow$) and visual style ($\downarrow$)}
            \label{fig:view+z_angle}
        \end{minipage}
        \hfill
        \begin{minipage}[b]{0.475\textwidth}  
            \centering 
            \includegraphics[width=\textwidth]{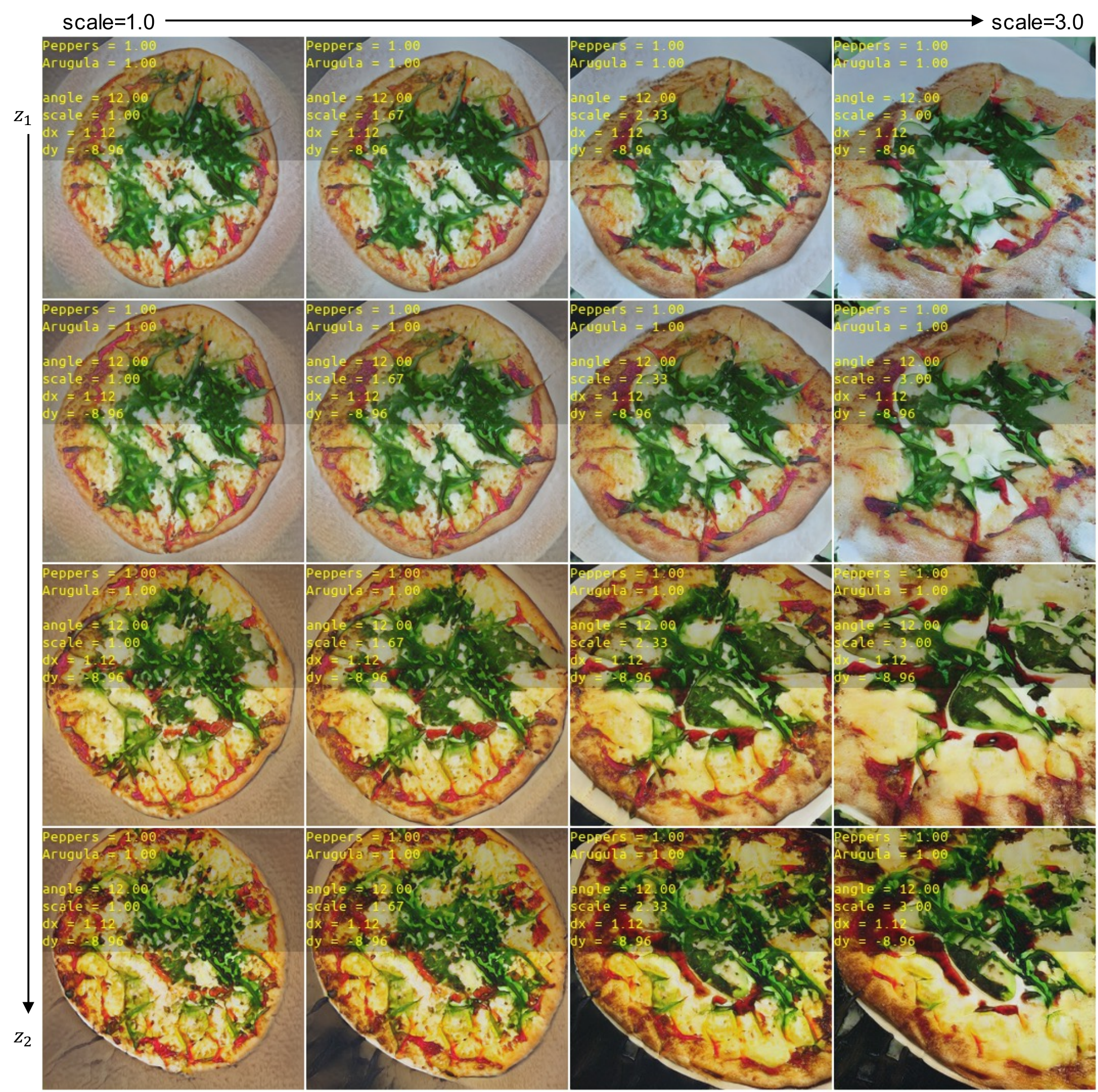}
            \vspace{0.2em}
            \caption*%
            {(b) scale ($\rightarrow$) and visual style ($\downarrow$)}
            \label{fig:view+z_scale}
        \end{minipage}
        \vskip\baselineskip
        \begin{minipage}[b]{0.475\textwidth}   
            \centering 
            \includegraphics[width=\textwidth]{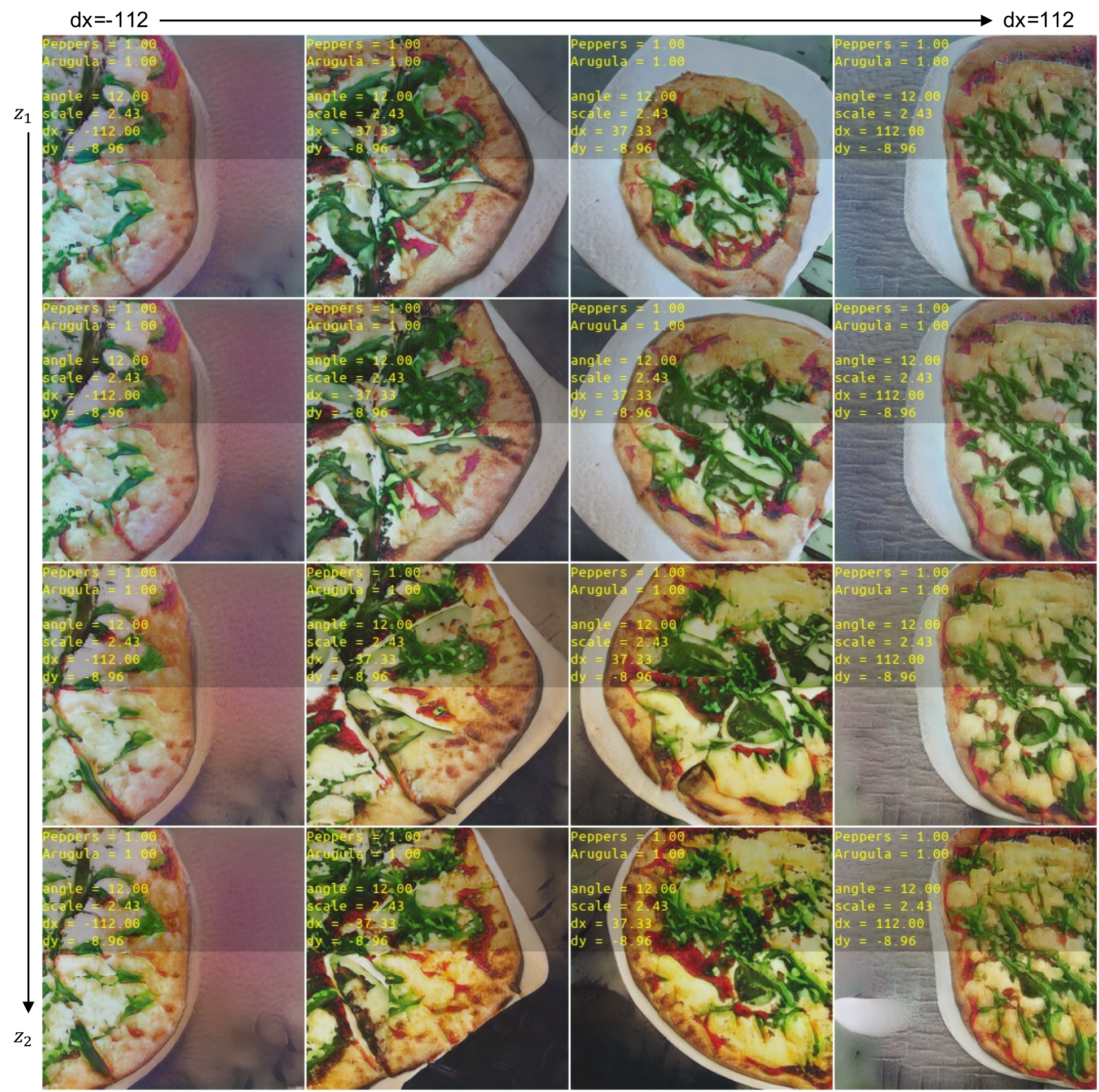}
            \vspace{0.2em}
            \caption*%
            {(c) dx ($\rightarrow$) and visual style ($\downarrow$)}
            \label{fig:view+z_dx}
        \end{minipage}
        \hfill
        \begin{minipage}[b]{0.475\textwidth}   
            \centering 
            \includegraphics[width=\textwidth]{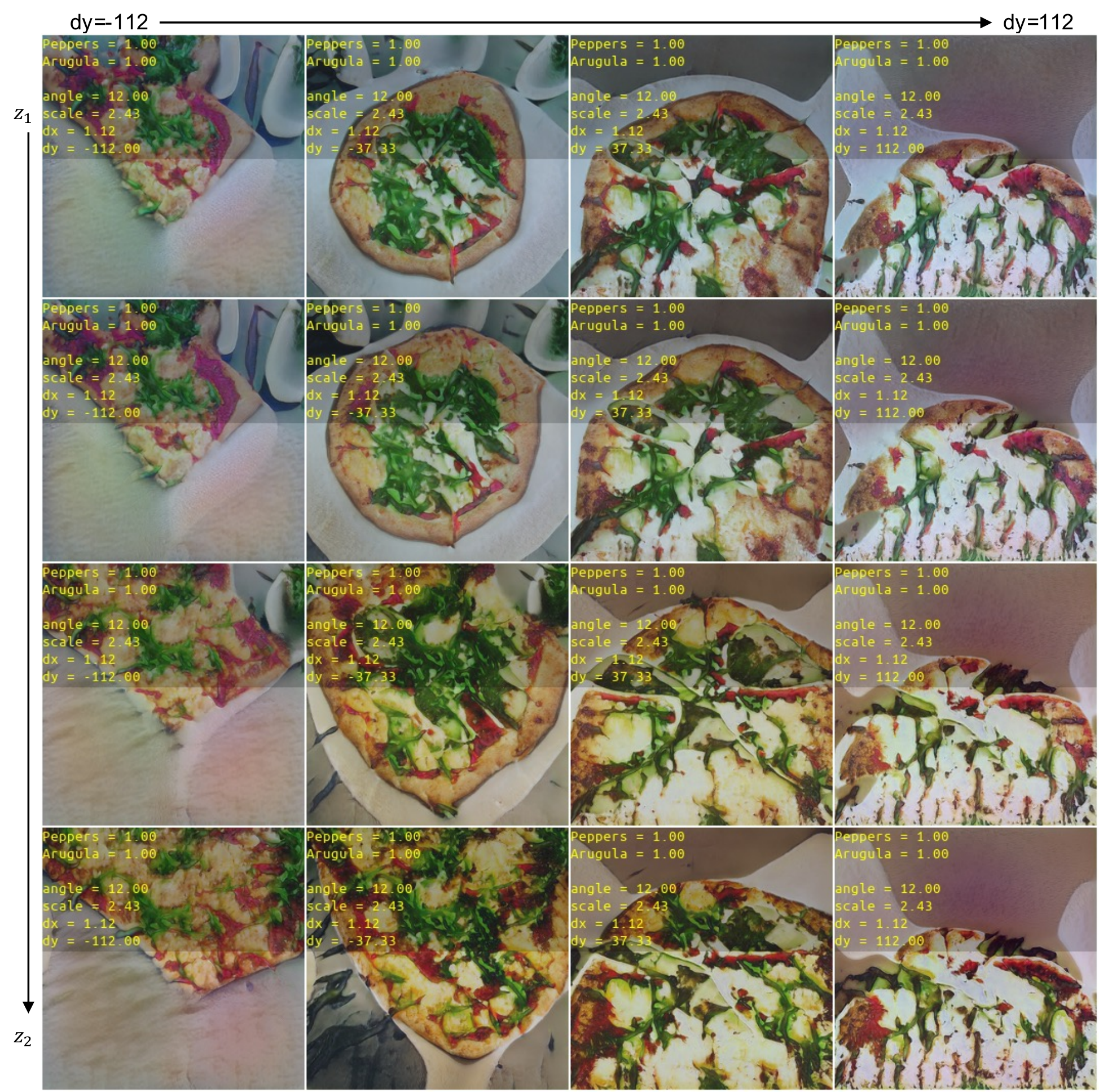}
            \vspace{0.2em}
            \caption*%
            {(d) dy ($\rightarrow$) and visual style ($\downarrow$)}
            \label{fig:view+z_dy}
        \end{minipage}
        \vspace{1em}
        \caption
        {Images generated using our \MPG by fixing the ingredients and traversing through \textbf{one view attribute} (along horizontal axis) and \textbf{visual styles} (along vertical axis). Notice that the ingredients are consistent and constant as we change the view attributes and the visual style. The visual style can be retained when changing the view attributes, \eg the cutting marks from $z_1$ appear at the first row of each sub-figure and disappear as we move to $z_2$.}
        \label{fig:fix_ingr}
    \end{figure*}

\begin{itemize}
    \item \autoref{fig:traversing_attributes}: Fix visual style and traverse ingredients and view attributes.
    \item \autoref{fig:ingr+z}: Fix view attributes and traverse ingredients and visual style.
    \item \autoref{fig:fix_ingr}: Fix ingredients and traverse view attributes and visual style.
\end{itemize}





\section{Dependency and Disentanglement of Attributes}
\label{sec:independence_between_attributes}
We seek to quantitatively verify the ability of \MPG to independently and effectively control the sets of attributes in synthesized images.  To do so, we perform a correlation analysis among all pairs of attributes in images synthesized by \MPG.  Namely, we first select one attribute as the controlling (input) attribute and then examine (a) whether that attribute is correctly depicted in the image and (b) whether all other attributes are not affected by the controlling attribute.  As the proxy for the assessed attribute, we use the output of the learned regressor / classifier for that attribute.  

\begin{figure*}[t!]
    \centering
    \includegraphics[width=\textwidth]{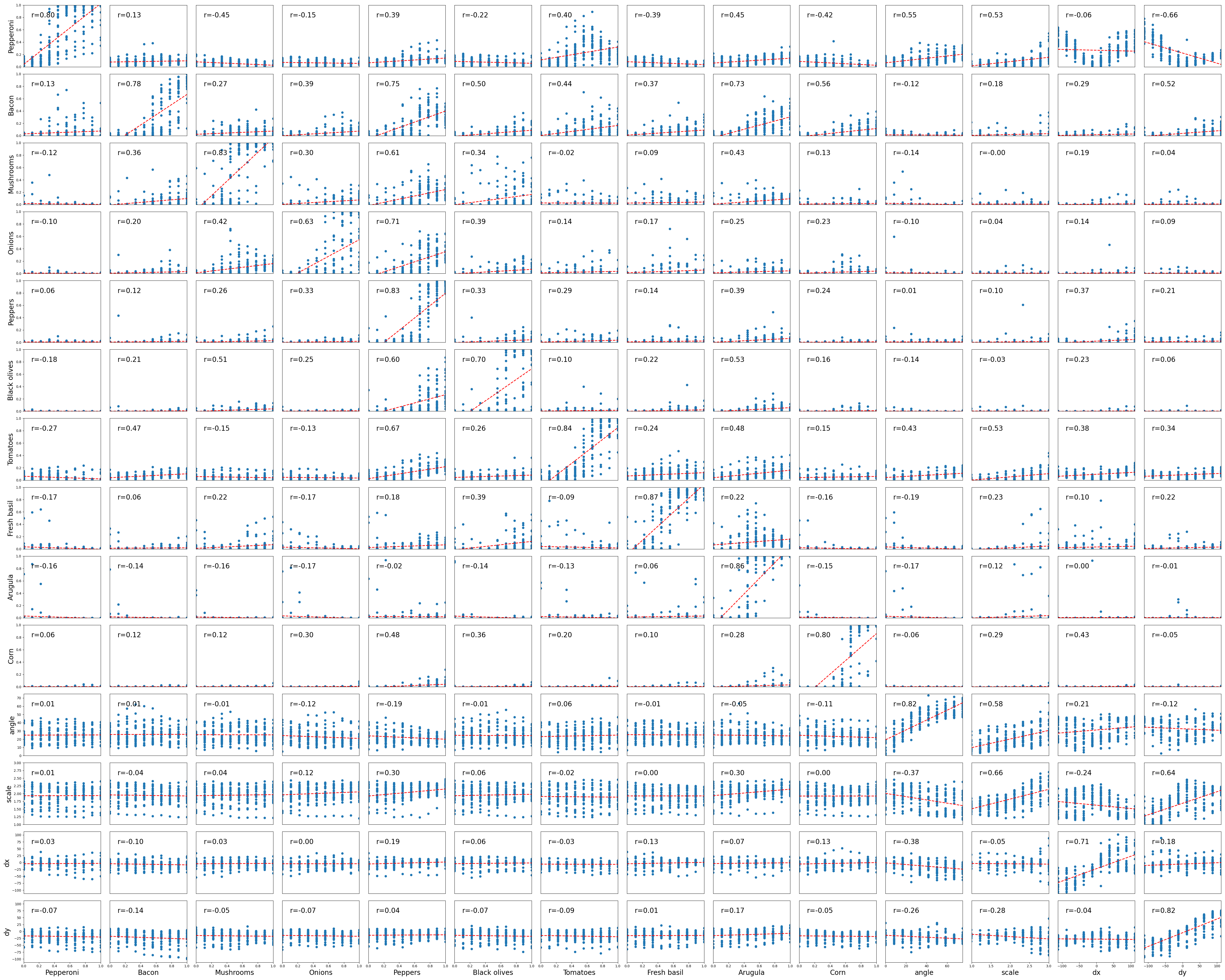}
    \vspace{0.2em}
    \caption{Dependency analysis of the predictions of an attribute, from synthesized images, (vertical axis in each plot) as a function of the controlling input attribute to \MPG (horizontal axis of each plot).  Each plot corresponds to a pair of (controlling input, predicted output) attributes.  For instance, the top row of plots has the Pepperoni attribute as the predicted output, while the last column of plots has the vertical shift (dy) as the controlling input.  Each plot also lists the correlation between the input and the output in the upper left corner.  The red lines show the linear input-output trends (best fit).  As expected, the diagonal plots, where the controlling input and the predicted output correspond to the same attribute, show strong correlation (diagonal trend), indicating the ability to control those attributes in the synthetic images.  On the other hand, the off-diagonal plots show significantly lower input-output correlation (uniform scatter).  This indicates that the predicted attribute is disentangled from (independent of) the controlling input.  See \autoref{sec:independence_between_attributes} for more details.}
    \label{fig:correlations}
\end{figure*}

\autoref{fig:correlations} depicts the results of this analysis. Each column corresponds to the traversal of one specific controlling (input) attribute; each row corresponds to the choice of a predicted attribute.  For example, to plot the prediction of Bacon in synthetic images when controlling Pepperoni as the input attribute (row 2, column 1), we gradually increase Pepperoni input $x_i$ from zero to one (ingredient labels are treated as continuous values as in the main paper). At each input value $x_i$, we generate 24 synthetic images and gather the corresponding Bacon ingredient classifier outputs $\{ y_{i}^{(1)}, y_{i}^{(2)}, ...  , y_{i}^{(24)} \}$.  Blue points in each plot are the the scatter plot of $x$ vs. $y$. After traversing through all $x_i$s, we also compute the correlation coefficient between the controlling attribute `Pepperoni' $x$ and the predicted outputs `Bacon' $y$ and plot the trend line.

Focusing on column one (i.e., controlling Pepperoni), we can observe that as we increase the Pepperoni value in the \MPG input (the "amount" of pepperoni), the prediction of Pepperoni from the synthetic image increases (row 1, column 1), indicating our ability to effectively control Pepperoni attribute.  We also notice the prediction of Bacon (row 2, column 1) is almost a flat line (uniform scatter), which indicates that the existence of Bacon in \MPG image is not influenced by the changing Pepperoni value in the input.  Similar flat lines / uniform scatter also appear across other rows in column 1, suggesting that Pepperoni attribute is independent of and disentangled from other ingredients as well as the view attributes.

\begin{figure}[ht]
    \centering
    \includegraphics[width=0.48\textwidth]{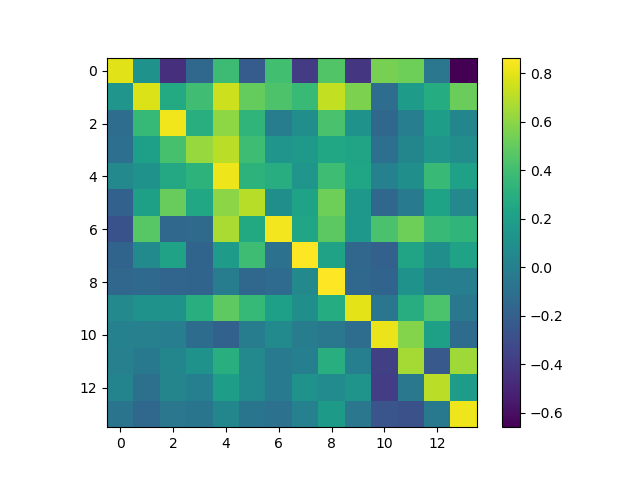}
    \caption{Heat map of correlation coefficients between each pair of (controlling input, predicted output) attributes, corresponding to numeric values of the correlation in the plots of \autoref{fig:correlations}.  The diagonal displays stronger correlation, as desired, from that in the off-diagonal pairs}
    \label{fig:coef_map}
\end{figure}

The same observation holds for other (control,predict) pairs of ingredients and view attributes. The off-diagonal plots generally have smaller correlation coefficients compared with diagonal ones (see also the correlation heat map in \autoref{fig:coef_map}), demonstrating the independence between attributes and the ability to control them effectively. 


\section{Comparison with SeFa on Attribute Control}

\begin{figure*}[t]
    \centering
    \includegraphics[width=0.96\textwidth]{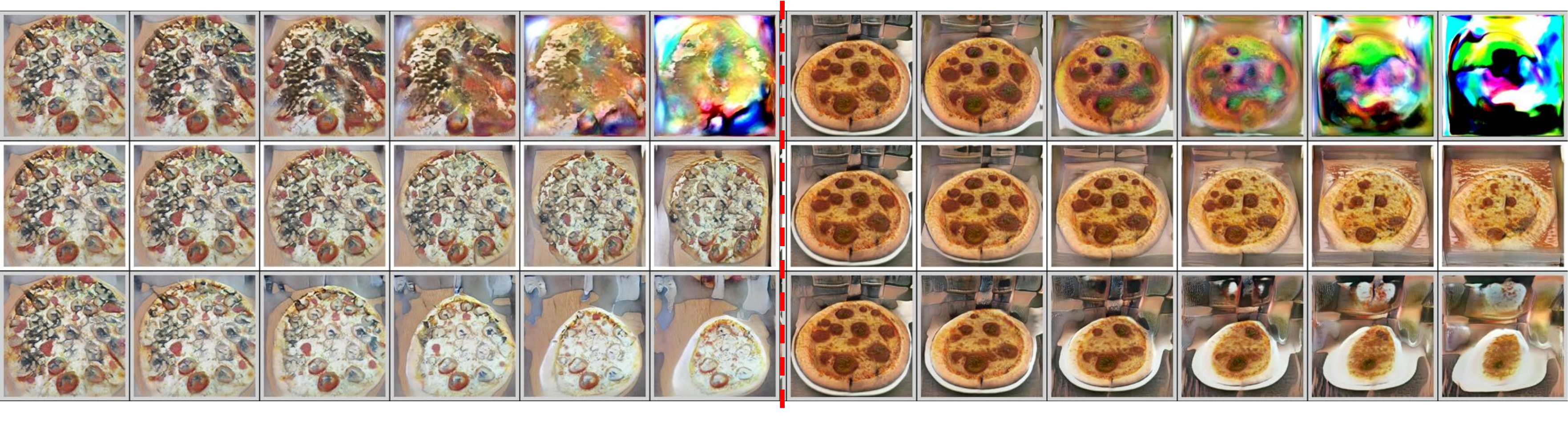}
    \vspace{0.5em}
    \caption{Two examples from SeFa~\cite{shen2020closed}. For each example, three largest eigen values as well as images moving along the corresponding eigen vectors are displayed (from top to bottom)}
    \label{fig:sefa}
\end{figure*}

To verify \MPG's ability to control ingredients and view attributes, we implement SeFa~\cite{shen2020closed} using pretrained \MPG model in \autoref{fig:sefa}. 
SeFa is an eigen-decomposition-based method for finding the most significant attribute directions in unconditional GAN models. 
We clone their official code\footnote{https://github.com/genforce/sefa}, load our pretrained \MPG and visualize the result in \autoref{fig:sefa}. The source images (i.e., the first column of each example) are generated from random ingredients and commonly seen view attributes, followed by traversal along the eigen-directions corresponding to the three largest eigen values of `all' layers (which performs better than those of the bottom, middle and top layers; refer to \cite{shen2020closed} for more details).
The limitation of SeFa is threefold: (1) SeFa needs to manually correlate the desired attributes with the eigenvectors to infer the direction's meaning; (2) The largest eigenvectors do not always correspond to a single attribute, because of entanglement; (3) We are unable to find the eigenvectors that control the ingredient appearance.

\section{Assessment of Images with Uncommon View Attributes}

\begin{figure}[ht]
    \centering
    \includegraphics[width=0.48\textwidth]{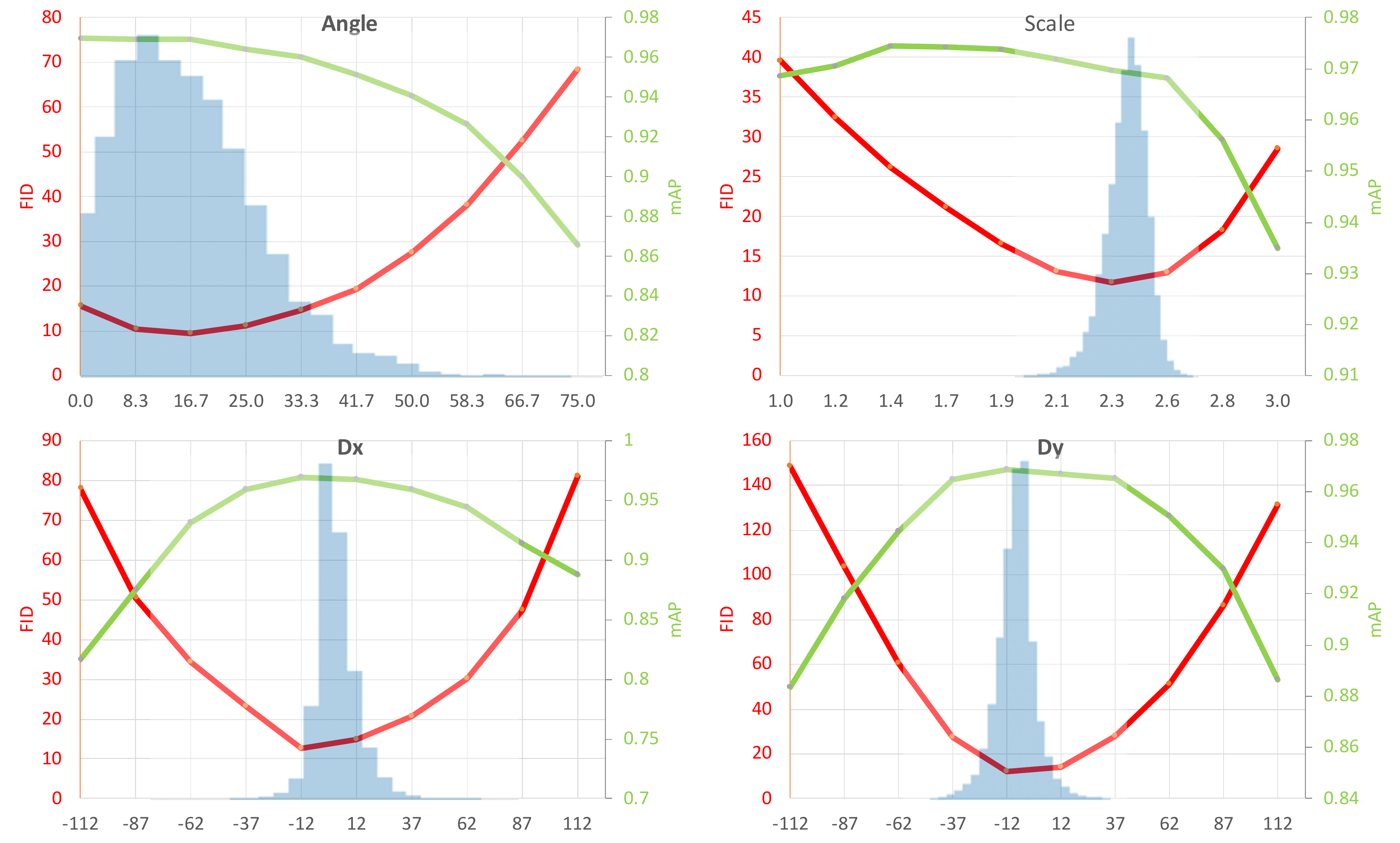}
    \vspace{0.5em}
    \caption{FIDs$\downarrow$ and mAPs$\uparrow$ as a function of the conditioning view attribute (angle, scale, dx, and dy). FIDs is estimated from 5k fake samples and all real images in \pizzaIngr; mAP is estimated using 5k fake samples. The histograms depict the distribution of the predicted attribute values on \pizzaIngr dataset}
    \label{fig:conditional_FID_and_mAP}
\end{figure}

We propose conditional FID and conditional mAP (i.e., c-FID and c-mAP) to verify the \MPG's performance at different view attribute values. 
To understand how c-FID and c-mAP are computed, we plot the predicted attribute value distributions along with the FIDs and mAPs of ingredients at different attribute values in \autoref{fig:conditional_FID_and_mAP}.
Taking the angle attribute as an example: the histogram is estimated using all samples in \pizzaIngr; to compute FID and mAP at one specific angle value are computed, we fix the angle value and randomly sample other view attributes and ingredients to generate 5k fake images to estimate the fake image distribution and mAP, followed by the FID between the \textit{fake distribution@5k} and \textit{real distribution@\pizzaIngr}. We observe that for each attribute, more frequent attribute values often result in improved FID and mAP,  leading to better image quality and ingredients control. 

c-FID (respectively c-mAP) of one attribute is computed by taking the average of the FIDs (respectively mAPs) at 10 evenly spaced values in the \textbf{targeting range} of the attribute. 
Since the estimated view attributes of real images in \pizzaIngr are not uniformly distributed in the targeting range as shown in the frequency histograms in \autoref{fig:conditional_FID_and_mAP}, we also compute c-FIDs and c-mAPs within three standard deviations around the mean value (i.e., \textbf{predicted range}) for each attribute (shown as \MPG$_{3\sigma}$).
The mean and standard deviation are estimated by fitting a Gaussian distribution from the frequency histogram.

We compare c-FID between \MPG and its counterparts in \autoref{tab:cFID}. \MPG exhibits better c-FID compared with those without \MSMAE, which again affirms the effectiveness of MSMAE.
Note that \MPG-\AR has the best c-FID; this is because without attribute regularizers, the model lacks sensitivity to view labels, able to generate reasonable images even at extreme view attribute values.
Comparing c-mAP in \autoref{tab:cmAP}, we  observe that removing \MSMAE again decrease c-mAPs.
Both c-FID and c-mAP are further improved by setting the range to be the \textbf{predicted range}.

\begin{table}[t]
    \small
    \centering
    \begin{tabular}{cccccc}
        \toprule
        Models & c-FID$_{angle}$ & c-FID$_{scale}$ & c-FID$_{dx}$ & c-FID$_{dy}$ & avg c-FID\\
        \midrule
        \MPG-\MSMAE & $28.18$ & $22.74$ & $39.04$ & $67.20$ & $39.29$\\
        \MPG-\MSMAE* & $27.95$ & $22.89$ & $40.11$ & $68.63$ & $39.90$\\
        \MPG-\AR  & $\mathbf{16.19}$ & $\mathbf{12.25}$ & $\mathbf{12.96}$ & $\mathbf{13.12}$ & $\mathbf{13.63}$\\
        \MPG  & $26.75$ & $22.07$  & $39.26$ & $66.34$ & $38.61$\\
        \MPG$_{3\sigma}$ & $20.21$ & $15.98$  & $14.56$ & $15.39$ & $16.54$\\
        \bottomrule
    \end{tabular}
    \vspace{1em}
    \caption{c-FID comparison between \MPG and its counterparts with missing components, best values are depicted in \textbf{bold}}
    \label{tab:cFID}
\end{table}

\begin{table}[t]
    \small
    \centering
    \begin{tabular}{cccccc}
        \toprule
        Models & c-mAP$_{angle}$ & c-mAP$_{scale}$ & c-mAP$_{dx}$ & c-mAP$_{dy}$ & avg c-mAP\\
        \midrule
        \MPG-\MSMAE & $0.9378$ & $0.9611$ & $0.9232$ & $0.9204$ & $0.9356$\\
        \MPG-\MSMAE* & $0.9496$ & $0.9441$ & $0.9133$ & $0.9280$ & $0.9338$\\
        \MPG-\AR  & $0.2043$ & $0.1907$ & $0.2161$ & $0.2166$ & $0.2069$\\
        \MPG  & $0.9416$ & $\mathbf{0.9663}$ & $0.9221$ & $0.9377$ & $0.9419$\\
        \MPG$_{3\sigma}$ & $\mathbf{0.9654}$ & $0.9651$ & $\mathbf{0.9659}$ & $\mathbf{0.9664}$ & $\mathbf{0.9657}$\\
        \bottomrule
    \end{tabular}
    \vspace{1em}
    \caption{c-mAP comparison between \MPG and its counterparts with missing components, best values are depicted in \textbf{bold}}
    \label{tab:cmAP}
\end{table}



\end{document}